\documentclass[twocolumn,a4,10pt]{article}

\usepackage[a4paper, total={7in, 8in}]{geometry}

\usepackage{graphicx}%
\usepackage{multirow}%
\usepackage{amsmath,amssymb,amsfonts}%
\usepackage{amsthm}%
\usepackage{mathrsfs}%
\usepackage[title]{appendix}%
\usepackage{xcolor}%
\usepackage{textcomp}%
\usepackage{manyfoot}%
\usepackage{booktabs}%
\usepackage{algorithm}%
\usepackage{algorithmicx}%
\usepackage{algpseudocode}%
\usepackage{listings}%
\usepackage{hyperref}
\usepackage[capitalise]{cleveref}
\usepackage{url}
\usepackage{hhline}
\usepackage{multirow}
\usepackage{tcolorbox}
\usepackage{authblk}
\usepackage{soul}
\definecolor{mygray}{gray}{0.6}
\definecolor{fuchsia}{rgb}{0.57, 0.36, 0.51}
\definecolor{applegreen}{rgb}{0.55, 0.71, 0.0}
\definecolor{amethyst}{rgb}{0.6, 0.4, 0.8}
\newcommand{\Q}[1]{\noindent \textbf{Comment:}}

\newcommand{\revise}[1]{ \textcolor{black}{#1}}
\begin{document}

\title{Effectively Leveraging CLIP for Generating Situational Summaries of Images and Videos}

\author[1]{Dhruv Verma}

\author[2]{Debaditya Roy}

\author[1,2]{Basura Fernando}

\affil[1]{Centre for Frontier AI Research Singapore, Agency for Science, Technology and Research Singapore}

\affil[2]{Institute of High-Performance Computing, Agency for Science, Technology and Research Singapore}

\date{}

\maketitle

\begin{abstract}

Situation recognition refers to the ability of an agent to identify and understand various situations or contexts based on available information and sensory inputs. It involves the cognitive process of interpreting data from the environment to determine what is happening, what factors are involved, and what actions caused those situations. This interpretation of situations is formulated as a semantic role labeling problem in computer vision-based situation recognition.
Situations depicted in images and videos hold pivotal information, essential for various applications like image and video captioning, multimedia retrieval, autonomous systems and event monitoring. 
However, existing methods often struggle with ambiguity and lack of context in generating meaningful and accurate predictions. 
Leveraging multimodal models such as CLIP, we propose ClipSitu, which sidesteps the need for full  fine-tuning and achieves state-of-the-art results in situation recognition and localization tasks.
ClipSitu harnesses CLIP-based image, verb, and role embeddings to predict nouns fulfilling all the roles associated with a verb, providing a comprehensive understanding of depicted scenarios. 
Through a cross-attention Transformer, ClipSitu XTF enhances the connection between semantic role queries and visual token representations, leading to superior performance in situation recognition. 
We also propose a verb-wise role prediction model with near-perfect accuracy to create an end-to-end framework for producing situational summaries for out-of-domain images.
We show that situational summaries empower our ClipSitu models to produce structured descriptions with reduced ambiguity compared to generic captions.
Finally, we extend ClipSitu to video situation recognition to showcase its versatility and produce comparable performance to state-of-the-art methods. 
In summary, ClipSitu offers a robust solution to the challenge of semantic role labeling providing a way for structured understanding of visual media.  
ClipSitu advances the state-of-the-art in situation recognition, paving the way for a more nuanced and contextually relevant understanding of visual content that potentially could derive meaningful insights about the environment that agents observe.
Code is available at \url{https://github.com/LUNAProject22/CLIPSitu-video}.
\end{abstract}

\section{Introduction}
\label{sec:intro}
\begin{figure*}[t]
    \includegraphics[width=\linewidth]{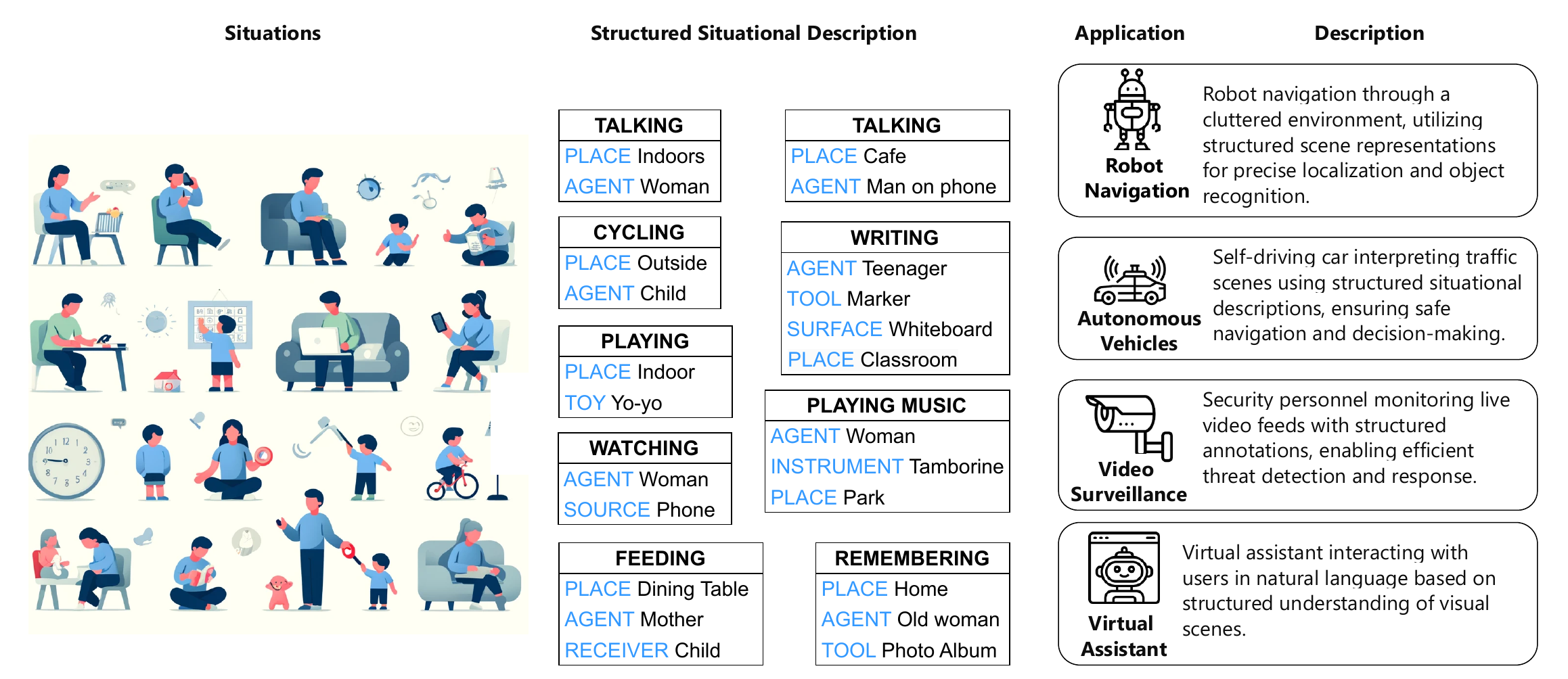}
    \caption{\textbf{Visual Understanding with Structured Situational Descriptions.} Each situation structure contains a verb in \textbf{bold} followed by the roles in \textcolor{blue}{blue} and the noun fulfilling that role. Structured situational descriptions provide a detailed and organized view of visual scenes, empowering assistive systems to understand, reason about, and interact with the world more effectively. From robot navigation \cite{liu2021robot}, autonomous vehicles \cite{malla2020titan}, video surveillance \cite{yuan2024towards,vishwakarma2013survey} to virtual assistants \cite{chiba2021dialogue}, structured representations enable a wide range of applications across diverse domains.}
    \label{fig:main-intro}
\end{figure*}

Accurately understanding the context and activities shown in visual media is essential for a wide array of applications. 
Vision based situation recognition defined in \cite{yatskar2016situation} stands at the intersection of computer vision and natural language processing and aims to identify and understand the activities, interactions, and scenarios from images and videos (see examples in \cref{fig:main-intro}). 
Situations as a representation of actions and their effects have been proposed way back by \cite{mccarthy1963situations, mccarthy1981some} in \textit{situational calculus}.  
They use situations to model the world at different points in time, actions to change these situations, and a set of axioms to describe how these actions transform one situation into another~\cite{reiter2001knowledge}. 
In event calculus proposed by \cite{kowalski1986logic}, a situation represents a snapshot of the world at a particular point in time. 
It captures the state of the world, including the events that have occurred up to that point and their effects. 
By capturing the dependencies and interactions among various entities (such as humans, objects, and scenes), situational description offers a rich depiction of the context in which actions unfold.
These structured situational descriptions can contribute to many applications shown in \cref{fig:main-intro} such as \textit{robot navigation} in a cluttered environment utilizing
structured scene representations for precise localization and detection of objects, \textit{autonomous vehicle} interpreting traffic scenes using structural description for safe navigation and decision making, \textit{video surveillance} with structured scene annotations for threat detection and response, and \textit{virtual assistant} interacting with users based on structured understanding of scenes.

Situational descriptions produce \textit{action centric representation} that describe the main action and various interactions between humans and objects in that action which can be evaluated for correctness precisely.
Furthermore, situational descriptions are \textit{richer representation} enables more nuanced understanding of scenes, including the relationships between objects, spatial arrangements, and temporal dynamics \cite{gu2019scene}.
Additionally, the structured nature of situational descriptions lends itself well to reasoning and decision-making tasks in assistive systems. 
By explicitly representing objects, actions, and relationships in a structured format, these descriptions enable more sophisticated reasoning about scenes and support complex decision-making processes \cite{zhu2017visual, puig2020watch, wan2022handmethat}.
Similarly, structured situational descriptions are easy to integrate with external knowledge bases, allowing AI systems to leverage additional information about objects, events, and concepts \cite{lu2016visual}. 
This integration enhances the understanding and interpretation of visual scenes, particularly in domains where external knowledge is crucial, such as in medical imaging or scientific research.
Finally, structured situational descriptions enable multi-modal reasoning by providing a unified framework for integrating visual information with other modalities, such as text or knowledge bases. 
This capability is essential for tasks that require understanding across multiple modalities, such as visual question answering or image-text retrieval \cite{yu2018mattnet}.


With the need for situational description established, we now describe \textbf{situation recognition} as per \cite{yatskar2016situation}.
Situation recognition comprises of verb prediction that describes the main activity in the image and semantic role labeling, which involves assigning semantic roles to different elements of the scene or action being executed--see~\cref{fig:main-intro}.
Semantic role labeling presents a several challenges, primarily due to the inherent ambiguity and variability in language and visual context. 
For instance, a single activity verb like "talking" expresses a spectrum of functional meanings and purposes, depending on the specific context captured in the image or video. 
Consider scenarios where ``talking" might involve a person talking to another person, talking on the phone, or talking on the stage -- see \cref{fig:main-intro}.
Therefore, effective semantic role labeling necessitates a fine-grained understanding of the event, leveraging contextual cues from both the visual content and linguistic definitions associated with the activity and its specific roles.

\textbf{Multimodal models for Situation Recognition.} Multimodal models like CLIP \cite{radford2021learning}, ALIGN \cite{jia2021scaling}, Flamingo \cite{alayrac2022flamingo}, VILA \cite{Lin_2024_CVPR}, and LLaVa \cite{liu2024llavanext} have been pivotal in developing a joint semantic representation of vision and language.
Trained on millions of image/text pairs, these models excel in capturing cross-modal dependencies between images and text. 
For instance, CLIP encounters various usages of verbs across visually diverse yet semantically similar images, providing a robust foundation for image semantic role labeling tasks. 
While CLIP demonstrates prowess in understanding both visual and linguistic information \cite{doveh2023teaching}, other approaches leverage multimodal models differently. 
Methods like VL-Adapter \cite{sung2022vl}, AIM \cite{yangaim}, EVL \cite{lin2022frozen}, and wise-ft \cite{wortsman2022robust} apply MLPs on top of image encoders for tasks like image classification or action detection. 
However, semantic role labeling require a different approach, requiring both verb and role alongside the image.
In addressing this challenge, Li et al. \cite{li2022clip} propose CLIP-Event, a fine-tuned CLIP model for situation recognition via text-prompt-based prediction. 
Despite leveraging the vast world knowledge of GPT-3, CLIP-Event falls short in semantic role labeling compared to CoFormer \cite{cho2022collaborative}, which is directly trained on images. 

We present \textbf{ClipSitu}, a novel approach for semantic role labeling without the need for fine-tuning the CLIP model. 
We leverage CLIP-based image, verb and role embeddings and predict all roles for a verb by sharing information across them.
We propose a cross-attention Transformer to learn the relation between semantic role queries and CLIP-based visual token representations.
We term this model as ClipSitu
XTF and it obtains state-of-the-art results for Situation Recognition on imSitu dataset \cite{yatskar2016situation}.
We also introduce a new MLP model that uses the cross-attention scores from ClipSitu XTF to localize the noun for a role in the image and obtain state-of-the-art results for noun localization.
We show that \textit{noun localization using cross-attention scores} is more effective than existing approaches even when using predicted verbs which demonstrates the robustness of our XTF cross-attention mechanism.

We presented image situation recognition using ClipSitu at WACV 2024 \cite{roy2024clipsitu}. 
We build upon \cite{roy2024clipsitu} and present the following additional \textbf{contributions}.
First, for video situation recognition \cite{sadhu2021visual}, we extend ClipSitu models (MLP, TF and XTF) to accept CLIP image embeddings of multiple frames from a video.
Then, we attach a transformer decoder to generate nouns as text phrases from the output of the ClipSitu models.
We show that with these extensions, ClipSitu framework produces comparable performance to state-of-the-art on video situation recognition -- see \revise{\cref{sec:vidsitu_revised} for the model description and \cref{sec:vidsitu_abl,sec:vidsitu_impl,sec:vidsitu_sota} for the results.}
Second, we evaluate a state-of-the-art Large Visual Language Model (VLM) on the structured prediction task of semantic role labeling with in-context examples \cref{sec:lvlm}. 
On both image and video semantic roles labeling, we observe that semantic role labeling is still a challenging task for VLM with in-context examples.
Therefore, we fine-tune multiple VLMs in \cref{sec:lvlm} to obtain improvement on both image and video situation recognition.
However, we also show that task specific models such as our ClipSitu models perform reasonably well with a fraction of the parameters of the VLMs.
Third, existing semantic role labeling frameworks require the semantic role to be provided for noun prediction. 
This may be a limitation when someone wants to deploy the models in real word as for some situations the exact frame structure may not be known beforehand. 
Instead, we predict roles associated with a verb with near perfect accuracy using image CLIP features (\cref{sec:role_pred}).
Adding role prediction to ClipSitu allows us to create the first end-to-end framework for describing images with situational frames.
We leverage our end-to-end framework to produce situational summaries of out-of-domain images from Conceptual Captions \cite{sharma2018conceptual} -- see \cref{sec:captioning}.

i\section{Related Work}
\subsection{Situation Recognition}
Situation Recognition approaches can be broadly classified into two categories -- one-stage where the situational verb, and nouns for the semantic roles are predicted simultaneously and two-stage where the verb and nouns are predicted in succession.

\subsubsection{One-stage Prediction} 
Many existing approaches exist for one-stage approaches predict both situational verbs and the associated nouns simultaneously from images  \cite{yatskar2016situation,yatskar2017commonly, li2017situation, suhail2019mixture}.
These approaches eliminate the need for sequential prediction of verbs from images and then roles from the verb and image. 
One-stage prediction reduces computational complexity capturing relationships between roles, nouns, and verbs in a single pass.
One-stage models such as Gated Graph Neural Network (GGNN) \cite{li2017situation} capture complex and varying relationships between roles without strict sequential constraints. 
GGNNs can represent  non-linear dependencies between entities represented as nodes in a graph.
Other one-stage approaches such as Conditional Random Fields (CRF) \cite{yatskar2016situation} and tensor decomposition models \cite{yatskar2017commonly} consider global dependencies across all roles and nouns simultaneously, potentially leading to more coherent predictions.
CRFs are probabilistic graphical models used in  \cite{yatskar2016situation} to model dependencies between random variables i.e. verbs and roles. 
Tensor decomposition factorize tensors on top of CRF models to handle interactions between roles and nouns efficiently, improving model scalability and performance \cite{yatskar2017commonly}.
Another one-stage approach called mixture kernels \cite{suhail2019mixture} provides a prior distribution for predicting nouns based on identified relationships in the image. 
This approach enhances prediction robustness, especially for rare or underrepresented noun-role combinations.

One-stage approach for video situation recognition \cite{zhao2023constructing} leverages spatio-temporal scene graph (HostSG) to model fine-grained spatial semantics and temporal dynamics within and across events. 
HostSG bridges the gap between the underlying scene structure and high-level event semantics. They jointly predict the event verb, semantic roles, and event relations, avoiding the error propagation across tasks.
However, one-stage models require extensive training data to generalize well across diverse scenes and contexts as they need to jointly model the verb and roles directly from the images.
Models like CRFs and tensor decomposition struggle with capturing long-range dependencies across multiple roles and verbs.
This issue is also prevalent in other approaches \cite{mallya2017recurrent, pratt2020grounded} that predict nouns in a predefined sequential order.
A final limitation of one-stage prediction is that model becomes complex to integrate complex relationships between roles, nouns, and verbs in a single pass increases training time.

\subsubsection{Two-stage Prediction} 
Two-stage prediction approaches in situation recognition introduce an additional stage to refine verb predictions using predicted nouns \cite{cooray2020attention,cho2021gsrtr,cho2022collaborative,wei2022rethinking}. 
Two-stage approaches \cite{wei2022rethinking} also allow for refinement of verb predictions based on more accurate noun predictions to improve overall verb prediction accuracy.
Another major advantage of these two stage approaches is using transformers that capture dependencies across longer sequences to provide comprehensive understanding of relationships between roles, verbs, and nouns.
Two-stage approaches \cite{cooray2020attention, cho2021gsrtr,cho2022collaborative} leverage transformers to predict semantic roles using interdependent queries. 
Other two-stage methods \cite{jiang2023exploiting} explore pre-trained multimodal transformers i.e. CLIP encoder on the image to predict the situational verb in the first stage and detect the objects in the second stage. 
Video situation recognition models capture temporal dependencies and contextual information across frames  using two-stage approach as well. Most methods \cite{sadhu2021visual,khan2022grounded, xiao2022hierarchical} use the transformer encoder to learn event wise contextual features from spatio-temporal features such as SlowFast \cite{feichtenhofer2019slowfast}. 
The decoder is used to learn the role relations and generate the nouns.
Grounded two-stage approaches with object features \cite{khan2022grounded} and object state embedding (pixel level changes), object
motion-aware embedding (motion of objects across multiple frames) and argument interaction embedding (relative position of multiple objects) \cite{yang2023video} improve the precision of noun localization in videos which is crucial for accurately describing events over time.

Two-stage approaches also suffer from limitations such as running separate predictions for verbs and nouns can increase computation compared to one-stage approaches.
Inaccurate verb predictions in the first stage can lead to cascading errors in subsequent noun predictions, necessitating robust error-handling mechanisms.
In video situation recognition, computational complexity is higher compared to image-based approaches especially with the need to compute spatio-temporal features.
We rely on CLIP image feature derived video representation to train ClipSitu for video situation recognition.
We adopt a two-stage mechanism for image situation recognition in ClipSitu for its advantages but train the verb and noun prediction separately to avoid propagation of errors. 
We also propose efficient cross-attention transformers for noun prediction and cross-attention score based MLP for localization of nouns.
Similarly, for video situation recognition, we add a lightweight transformer decoder to ClipSitu for generation of role-wise nouns.

\subsection{Structured Representation for Captioning}
In the previous subsections, we have summarized approaches that focus on structured summary of images and videos using the situational framework. 
Structured representations have also been explored extensively for generating balanced image captions \cite{jia2023image,nguyen2021defense,yang2023transforming,ye2022hierarchical,gu2023text} and video captions \cite{li2023learning,zhao2023constructing,wang-etal-2019-role, yang2023transforming,wang-etal-2023-improving,lu2024set}.
The structured representation used by these approaches are scene graphs that encode relationships between objects (nouns) and actions (verbs), providing a structured representation of visual scenes 
\cite{jia2023image} and \cite{nguyen2021defense}.
\cite{wang-etal-2023-improving} propose Structured Concept Predictor (SCP) that enhances image captioning by predicting concepts and their relationships before generating captions.
Similarly for video captioning, the Set Prediction Guided by Semantic Concepts (SCG-SP) framework \cite{lu2024set} generates multiple captions for a single video by encoding semantic concepts detected in the frames. 
The object and relationships denoted in the scene graph reduces the semantic gap between visual (image/video) and text modalities.
Therefore, captions generated using SG are better grounded and more relevant to the visual content \cite{wang-etal-2019-role, yang2023transforming}.

A limitation of constructing and utilizing scene graphs for caption generation is that it requires accurate object detection, relationship tagging, and large-scale graph reasoning.
Relationship prediction in images and videos is relatively inaccurate \cite{cong2021spatial,kim2024groupwise,li2024scene}.
Another challenge is the integration of scene graphs into caption generation pipelines requires careful alignment of graph representations with model architectures and training objectives.
Therefore, we show that a structured summary of an image is in itself a compact and informative description of the image compared to the ground truth captions provided for the image. 
We leverage the trained ClipSitu on imSitu to generate the structured summary of out-of-domain images without the necessity of fine-tuning, paving the way for adoption of structured prediction approaches for image description.

\section{CLIPSitu Models and Training}
In this section, first, we present how we extract CLIP~\cite{radford2021learning} embedding (features) for situation recognition.
After that we present the verb prediction model. 
Then, we present three models for Situation Recognition using the CLIP embeddings.
Finally, we present a loss function that we use to train our models.

\subsection{Extracting CLIP embedding}
Every image $I$ has a situational action associated with it, denoted by a verb $v$. 
For this verb $v$, there is a set of semantic roles $R_v = \{r_1, r_2, \cdots, r_m\}$ each of which is played by an entity denoted by its noun $N = \{ n_1, n_2, \cdots, n_m\}$.
We use CLIP~\cite{radford2021learning} visual encoder $\psi_v()$, and the text encoder $\psi_t()$ to obtain representations for the image, verb, roles, and nouns denoted by $X_I$, $X_V$, $X_{R_v}$ and, $X_N$ respectively.
Here $X_{R_v} = \{ X_{r_1}, X_{r_2}, \cdots, X_{r_m}\}$ {for $m$ roles} and $X_N = \{ X_{n_1}, X_{n_2}, \cdots, X_{n_m} \}$ {for corresponding $m$ nouns} where $X_V = \psi_t(v)$, $X_{r_i} = \psi_t(r_i)$ and $X_{n_i} = \psi_t(n_i)$ are obtained using text encoder. Similarly, the $X_I = \psi_v(I)$ is obtained using vision encoder. Note that all representations $X_I, X_V, X_{r_i}$ and $X_{n_i}$ have the same dimensions.

\subsection{ClipSitu Verb MLP}
The first task in situation recognition is to predict the situational verb correctly from the image.
We design a simple MLP with CLIP embeddings of the image $X_I$ as input as follows:
\begin{equation}
    \hat{v} = \phi_{V}(X_I).
\end{equation}
where $\phi_V$ contains $l$ linear layers of a fixed dimension with ReLU activation to predict the situational verb.
Just before the final classifier, there is a Dropout layer with a 0.5 rate.
We call this model as ClipSitu Verb MLP and we train it with standard cross-entropy loss.

\subsection{ClipSitu MLP}
\label{sec.meth.cmlp}
\begin{figure}
    \centering
    \begin{tabular}{cc}
    \includegraphics[width=0.35\linewidth]{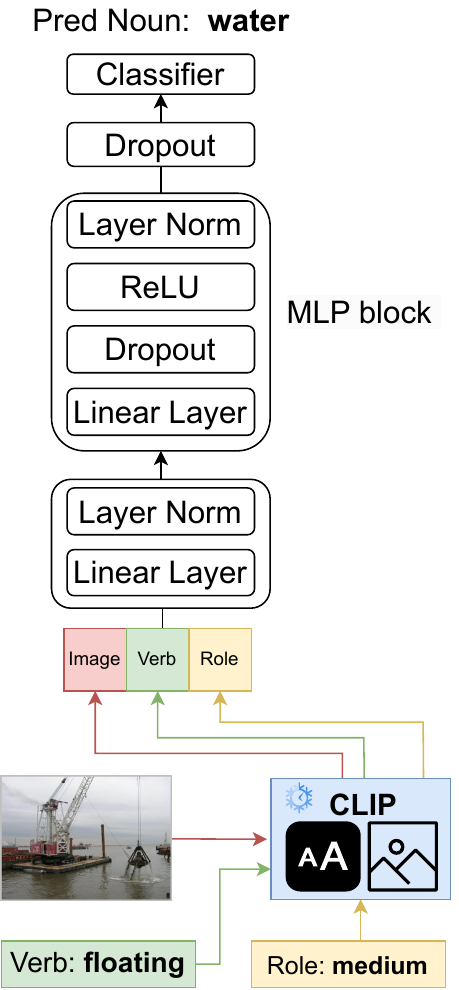} & \includegraphics[width=0.55\linewidth]{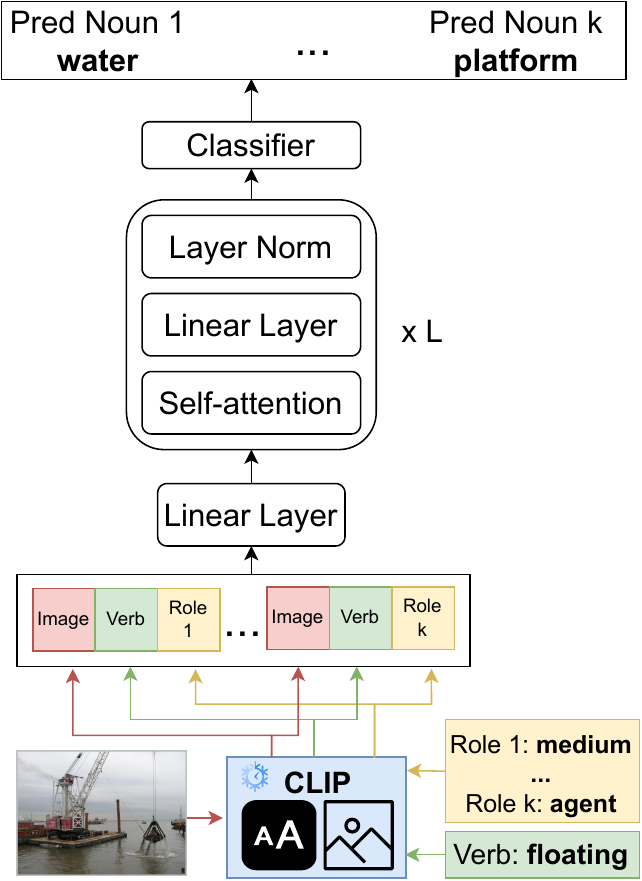} \\
        (a)     &  (b)  
    \end{tabular}
    \caption{{Architecture of (a) ClipSitu MLP and (b) ClipSitu TF. We use pooled image embedding from the CLIP image encoder for ClipSitu MLP and TF.}}
    \label{fig:clipsitu_mlp_tf}
\end{figure}

Next, we design a modern multimodal MLP block for semantic role labeling for Situation Recognition that predicts each noun of the semantic role associated with the predicted verb in a given image.
We term this method as \textbf{ClipSitu MLP}.
The detailed architecture of ClipSitu MLP is  shown in \cref{fig:clipsitu_mlp_tf}(a).
Specifically, given the image, verb, and role embeddings, the ClipSitu MLP predicts the embedding of the corresponding noun for the role.
In contrast to what has been done in the literature, ClipSitu MLP obtains contextual information by conditioning the information from the image, verb, and role embeddings.
While the image embedding provides context about the possible nouns for the role, the verb provides the context on how to interpret the image situation.

We concatenate the role embedding for each role ${r_i}$ to the image and verb embedding to form the multimodal input $X_i$ where $X_i = [X_I, X_v, X_{r_i}]$. 
Then, we stack $l$ MLP blocks to construct CLIPSitu MLP and use it to transform the multimodal input $X_i$ to predict the noun embedding $\hat{X}_{n_i}$ as follows:
\begin{equation}
    \hat{X}_{n_i} = \phi_{MLP}(X_i).
\end{equation}
In $\phi_{MLP}$, the first MLP block projects the input feature $X_i$ to a fixed hidden dimension using a linear projection layer followed by a LayerNorm ~\cite{ba2016layer}.
Each subsequent MLP block consists of a Linear layer followed by a Dropout layer (with a dropout rate of 0.2), ReLU~\cite{nair2010rectified}, and a LayerNorm.
We predict the noun class from the predicted noun embedding using a dropout layer (rate 0.5) followed by a linear layer which we name as classifier $\phi_c$ as 
\begin{equation}
    \hat{y}_{n_i} =\texttt{argmax}~ \phi_c(\hat{X}_{n_i})
\end{equation}
where $\hat{y}_{n_i}$ is the predicted noun class.
We use cross-entropy loss between predicted $\hat{y}_{n_i}$ and ground truth nouns ${y}_{n_i}$ as explained later in~\cref{loss} to train the model.

\subsection{ClipSitu TF: ClipSitu Transformer}\label{sec:tf}
The role-noun pairs associated with a verb in an image are related as they contribute to different aspects of the execution of the verb. 
Hence, we extend our ClipSitu MLP model using a Transformer~\cite{vaswani2017attention} to exploit the interconnected semantic roles and predict them in parallel.
The input to the Transformer is similar to ClipSitu MLP (i.e. $X_i = [X_I, X_v, X_{r_i}]$), however, we build a set of vectors using $\{ X_1, X_2, \cdots, X_m\}$ where $m$ denotes the number of roles of the verb.
Each vector in the set is further processed by a linear projection to reduce dimensions.
Our Transformer model $\phi_{TF}$ consists of $l$ encoder layers where each encoder layer have $h$ number of multi-head attention heads.
Using the Transformer model, we predict the noun embedding of the $m$ roles as output of the transformer
\begin{equation}
   \{ \hat{X}_{n_1}, \hat{X}_{n_2}, \cdots, \hat{X}_{n_m}\} = \phi_{TF}(\{ X_1, X_2, \cdots, X_m\}).
\end{equation}
Similar to the ClipSitu MLP (\cref{sec.meth.cmlp}), we predict the noun classes using a classifier on the noun embedding as $\hat{y}_i = \texttt{argmax}~ \phi_c(\hat{X}_{n_i})$ where $ i= \{1, \cdots, m\}$.
The ClipSitu Transformer model for short know as ClipSitu TF is shown in~\cref{fig:clipsitu_mlp_tf}(b).

\begin{figure}[t]
    \centering
    \includegraphics[width=0.9\linewidth]{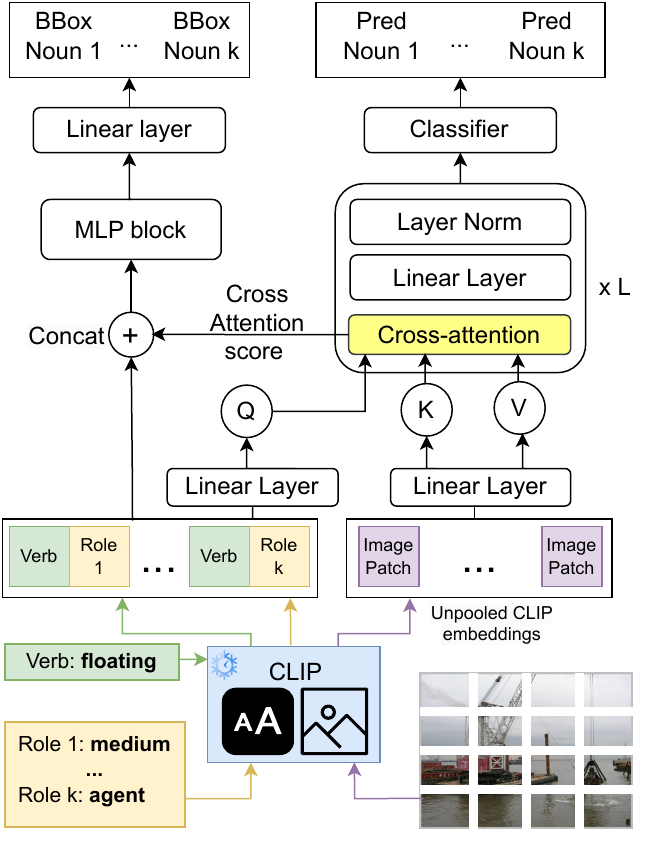}
    \caption{{Architecture of ClipSitu XTF. We use embeddings from each patch of the image obtained from CLIP image encoder.}}
    \label{fig:xtf}
\end{figure}

\subsection{ClipSitu XTF: Cross-Attention Transformer}
Each semantic role in a situation is played by an object located in a specific region of the image.
Therefore, it is important to pay attention to
the regions of the image which has a stronger relationship with the role.
Such a mechanism would allow us to obtain better noun prediction accuracy.
Hence, we propose to use the encoding for each patch of the image obtained from the CLIP model.
We design a cross-attention Transformer called \textbf{ClipSitu XTF} to model how each patch of the image is related to every role of the verb through attention as shown in \cref{fig:xtf}.

Let the patch embedding of an image be denoted by $X_{I,p} = \{ X^1_I, X^2_I, \cdots, X^p_I\}$ where $p$ is the number of image patches.
These patch embeddings form the key and values of the cross-attention Transformer while the verb-role embedding is the query in Transformer. 
The verb embedding is concatenated with each role embedding to form $m$ verb-role embeddings $X_{vr} = \{ [X_V; X_{r_1}], [X_V; X_{r_2}], \cdots, [X_V; X_{r_m}] \}$.
We project each verb-role embedding to the same dimension as the image patch embedding (i.e the embedding size of $X^p_I$) using a linear projection layer.
Then the cross-attention operator in a Transformer block is denoted as follows:
\begin{align}
Q = W_Q &X_{vr},~~ K = V = W_I X_{I,p} \nonumber\\
&\hat{X} = \text{softmax}\frac{QK^T}{\sqrt{d_K}} V
\end{align}
where $W_Q$ and $W_I$ represent projection weights for queries, keys, and values and $d_K$ is the dimension of the key token $K$. 
As with ClipSitu TF, we have $l$ cross-attention layers in ClipSitu XTF.   
The predicted output from the final cross-attention layer contains $m$ noun embeddings $\hat{X} =  \{ \hat{X}_{n_1}, \hat{X}_{n_2}, \cdots, \hat{X}_{n_m}\}$. 
Similar to the transformer in \cref{sec:tf}, we predict the noun classes using a classifier on the noun embeddings as $\hat{y}_i = \phi_c(\hat{X}_{n_i})$ where $ i= \{1, \cdots, m\}$.

{Next, we use ClipSitu XTF to perform localization of roles that requires predicting a bounding box $\textbf{b}_i$ for every role $r_i$ in the image.
The cross-attention matrix from the first layer of ClipSitu XTF is denoted by $A$ where $A \in \mathcal{R}^{m \times p}$ is rearranged into $m$ row vectors $\{A_1, \cdots, A_m\}$. 
Each row vector $A_i$ is $p$-dimensional and each element in the vector shows how each patch in the image is related to the verb-role pair. 
To incorporate the verb and role context to the score vector ($A_i$), we concatenate each score vector to its corresponding verb-role embedding from $X_{vr}$ to obtain input for localization denoted by $X_l$ as follows:
\begin{align}
    X_l =& \{[X_V; X_{r_1}; A_1], \\ \nonumber [X_V; & X_{r_2}; A_2], \cdots, [X_V; X_{r_m}; A_m] \}    
\end{align}

We pass $X_l$ through a single MLP block (designed for ClipSitu MLP) followed by a linear layer and a sigmoid function to obtain the predicted bounding box $\hat{b}_i \in [0, 1]^4$ for every role $r_i$. 
The four elements in the predicted bounding box indicate the center coordinates, height and width relative to the input image size. 
Though ClipSitu XTF uses cross-attention as CoFormer \cite{cho2022collaborative}, the verb role tokens in the query and the image tokens are obtained from CLIP and not learned.
Leveraging the power of CLIP embeddings allows us to design a simpler one-stage ClipSitu XTF compared to the two-stage CoFormer \cite{cho2022collaborative}.
}

\label{loss}
\textbf{Minimum Annotator Cross Entropy Loss.}  Popular situation recognition datasets employ multiple annotators to label each noun for a given role of an image instance. 
In some instances, annotators may not provide the same annotation for the same role as there can be multiple interpretations of the role.
Existing approaches \cite{cooray2020attention,cho2022collaborative} make multiple predictions instead of one to tackle this issue.
However, this can confuse the network during training as there are multiple annotations for the same example.
Furthermore, the loss function should not penalize a prediction that is close to any of the annotators' ground truth but further away from others. 
We propose minimum cross-entropy loss that considers the ground truth from each annotator. 
For a prediction $\hat{y}_i$, ground truth from all the annotators $\mathcal{O}= \{ O_1, \cdots O_q \}$ is used to train our network as follows
\begin{equation}
    \mathcal{L}_{MAXE} = \min_{\mathcal{O}} -\sum_{c=1}^C y^{(O_j)}_{i,c} \text{log}(\hat{y}_{i,c}) \text{ where } \forall O_j \in \mathcal{O}. 
\end{equation}
Here, $C$ denotes the total number of noun classes and $\mathcal{L}_{MAXE}$ stands for Minimum Annotator Cross Entropy Loss. 

To train ClipSitu XTF for localization of roles, we employ $L1$ loss to compare the predicted and ground-truth bounding boxes 
\begin{equation}
    \mathcal{L}_{L1} = \frac{1}{m}\sum^m_{i=1} \| b_i - \hat{b}_i \|_1.
\end{equation}
We train ClipSitu XTF for noun prediction and localization using the combined loss $\mathcal{L} = \mathcal{L}_{MAXE} + \mathcal{L}_{L1}$. 
\section{\revise{Extending ClipSitu for Video Situation Recognition}}
\label{sec:vidsitu_revised}

\revise{Video situation recognition \cite{sadhu2021visual} (VidSitu) introduces the task of describing the situation in a 10 second video. 
We need to predict a set of $k$ related events that constitute the situation $\{E_i\}_{i=1}^k$. 
Each event in the situation $E_i$ is described by a verb $v_i$ chosen from a set of verbs $V$ and the nouns (entities, location, or other details of the event) described in text that are assigned to various roles of the verb. 
While the concept of situation is similar in images and videos, there are key differences between how imSitu and VidSitu are constructed.
First, the number of verbs in VidSitu is thrice that that of imSitu (1560 vs. 504) as it distinguishes between homonyms based on verb-senses for e.g.  ``strike (hit)'' is different from ``strike (a pose)''.
Second, for each role in VidSitu, the noun is a text phrase instead of a noun class in imSitu e.g. ``man holding shield" vs. ``man".
The large vocabulary of verbs and noun phrases makes it challenging to predict the exact verb and noun. 
Therefore, situation recognition in VidSitu is set up as a generation task and evaluated using semantic similarity metrics instead of being a classification task (details in \cref{sec:vidsitu_eval_det}).
}

\begin{figure*}
    \includegraphics[width=\linewidth]{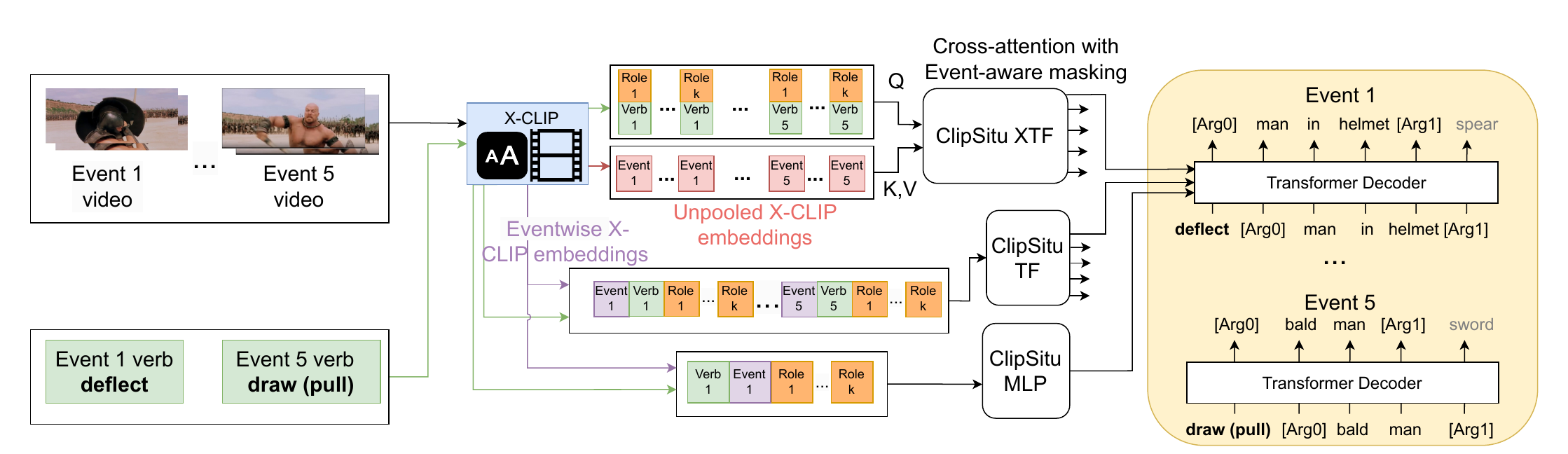}
    \caption{Extending ClipSitu models (XTF, TF, and MLP) for Video Situation Recognition (VidSitu). For every event video (2s duration) from a 10s video, X-CLIP \cite{ma2022x} provides the video embeddings. We use a transformer decoder (TxD) for generating the nouns playing different roles for the verb of an event. We generate nouns for all 5 events in a video together. For ClipSitu XTF, verb and role acts as query and unpooled X-CLIP embeddings as key and value. ClipSitu XTF uses \textit{event-aware masking} i.e. verb $i$ (+roles) tokens attend only to event $i$ tokens. For ClipSitu MLP, we collect the output of all 5 events before sending to the transformer decoder.
    }
    \label{fig:vidsitu_all_models}

\end{figure*}

\revise{\subsection{ClipSitu with decoder overview}}
\revise{ To extend ClipSitu models for videos and perform text generation, we make the following  changes to ClipSitu models. 
First, we use video embeddings such as X-CLIP\cite{ma2022x} for each event of duration 2 seconds in a 10 second video.
Second, we add a language decoder based on transformer to our ClipSitu models as shown in \cref{fig:vidsitu_all_models} as the noun prediction is a generation task in VidSitu instead of classification in imSitu.
The simultaneous noun prediction for each role in image situation recognition is replaced by sequential role-wise noun generation using the transformer decoder.
For the ClipSitu MLP and TF models, we use the coarse-grained (video to text) embedding from X-CLIP along with the verb embedding. 
For the ClipSitu XTF model, we use the fine-grained (image to text) unpooled video-embedding obtained from X-CLIP before temporal pooling similar to our strategy of using unpooled CLIP features for imSitu. 
We use verb and role embeddings obtained from X-CLIP as roles are unique for every verb in VidSitu.
In VidSitu \cite{sadhu2021visual}, roles are denoted by placeholders -- Arg0, Arg1, Arg2, ALoc (Location), AScn (Scene), ADir(Direction), and AMnr (Manner).
However, the actual roles depending on the verb which are also available in the VidSitu, for e.g. Arg0 for the verb \textit{driving} is \textit{driver} and for \textit{looking} is \textit{looker}.}

\revise{For all the ClipSitu models, the output obtained for each event video is fed as input to a transformer decoder to generate the nouns for each role. 
Apart from the ClipSitu outputs, the verb and ground-truth role and nouns are presented during training. 
The decoder is trained to generate a verb-noun sequence which is right shifted by one word as shown in \cref{fig:vidsitu_all_models}.
The placeholders Arg0, Arg1, Arg2, ArgScn are useful in evaluation as they are used to separate the generated nouns.
During testing, the decoder receives the ClipSitu outputs and generates the rolewise nouns.
VidSitu contains 5 events in a video that are semantically connected. 
Therefore, we also generate the nouns for all 5 events together which is shown to be effective in  existing works \cite{sadhu2021visual,khan2022grounded}.
For ClipSitu MLP, we collect the output of all 5 events before sending to the transformer decoder.
For ClipSitu XTF, verb and role act as query and unpooled X-CLIP embeddings as key and value. 
During cross-attention of XTF, event-aware masking is applied that restricts the verb and role of event $i$ to attend to only the unpooled X-CLIP features from event $i$.
The transformer decoder (TxD) that we add to our ClipSitu models is inspired by \cite{sadhu2021visual, xiao2022hierarchical}.
The encoder outputs from MLP, TF, and XTF are used as encoder attention to start the decoding. 
During training, cross-entropy loss is employed for each token in the predicted sequence.}

\revise{\subsection{ClipSitu with decoder details}}
\revise{For each event $E_i$ with video embedding $X_{E_i}$ (via X-CLIP), CLIP verb embedding $X_{v_i}$ and CLIP role embeddings for each role ${ X_{r_i}^{Arg0}, X_{r_i}^{Arg1}, X_{r_i}^{Arg2}, X_{r_i}^{ArgScn}}$, the input to TxD is:
\begin{equation}
    X^{(i)} = \phi_{model}([X_{E_i};X_{v_i};X_{r_i}]),
\end{equation}
where $\phi_{model}$ is either Clipsitu MLP, TF, or XTF, yielding embeddings for each role’s noun.}

\revise{\textbf{Event-aware masking.}
In TF and XTF models, if an event’s embeddings attempt to access frames from adjacent events, it could lead to information leakage across events, which would compromise the temporal coherence and accuracy of role-noun assignments within each event.
With event-aware masking, we create a binary mask $M_{i,j}$ such that:
$M_{i,j} = 1 \text{  if } i = j$ (event $E_i$ can attend only to itself) and $ M_{i,j} = 0 \text{  if }  i \ne j$ (no cross-event attention is allowed).
Let’s denote the query matrix for roles in event $E_i$ as $Q^{(i)}$ and the key and value matrices for frames in event $E_i$ as $K^{(i)}$ and $V^{(i)}$, respectively.
As the inputs from all events is fed together to the TxD, the attention score between query and key matrices across events is computed as
$A_{i,j} = \text{softmax}\frac{Q^{(i)} K^{(j)\top}}  {\sqrt{d_K}}$.   
The masked output is computed using the binary mask and the attention score
\begin{equation}
    \hat{X} = A_{i,j} \cdot M_{i,j} \cdot V_i
\end{equation}
By enforcing that each role-specific query only attends to its corresponding event’s frame embeddings, the model is better able to focus on the local visual context relevant to that event.
For instance, if event $E_1$ involves a "driving" action with the "driver" (Arg0) role, and event $E_2$ involves a "talking" action with a "speaker" (Arg0) role, event-aware masking prevents the "driver" role in $E_1$ from erroneously attending to frames in $E_2$.
This localized attention structure preserves the temporal structure of the video and ensures that each event’s role predictions are more contextually accurate.}

\revise{\textbf{Loss for video situation recognition.} 
For image situation recognition, cross-entropy is used for noun prediction	but for video situation recognition, cross-entropy loss is computed for sequence generation across multiple events.
The transformer decoder TxD takes in ClipSitu output $X^{(i)}$ as input and outputs a right-shifted role-noun sequence $\hat{y}^{(i)}$ for each event of length $T^{(i)}$.
The ground truth noun sequence is given by ${y}^{(i)} = X_{r_i}^{Arg0}, X_{r_i}^{Arg1}, X_{r_i}^{Arg2}, X_{r_i}^{ArgScn}$.
The cross-entropy loss is applied to the predicted noun sequence during training, calculated as:
\begin{equation}
\mathcal{L}^{(i)} = -\sum_{t=1}^{T^{(i)}} \sum_{c=1}^C {y}^{(i)}_{t,c} \text{log} \hat{y}^{(i)}_{t,c}
\end{equation}
where $C$ is the size of vocabulary. 
Ground truth and predicted noun sequences ($y^{(i)}$ and $\hat{y}^{(i)}$) are aligned by TxD.}

\section{ClipSitu Ablations}
\subsection{imSitu Evaluation Details}
We perform our experiments on imSitu dataset \cite{yatskar2016situation} and the augmented imSitu dataset with bounding boxes called SWiG \cite{pratt2020grounded} for the tasks of situation recognition and noun localization, respectively. 
Situation recognition with noun localization is called \textit{grounded situation recognition.}
The dataset has a total of 125k images with 75k train, 25k validation (dev set), and 25k test images.
The metrics used for semantic role labeling are \textit{value} and \textit{value-all} \cite{yatskar2016situation} which predict the accuracy of noun prediction for a given role.
For a given verb with $k$ roles, \textit{value} measures whether the predicted noun for at least one of $k$ roles is correct.
On the other hand, \textit{value-all} measures whether all the predicted nouns for all $k$ roles are correct. 
A prediction is correct if it matches the annotation of any one of the three annotators. 
{Noun localization metrics \textit{grnd value} and \textit{grnd value-all} compute the accuracy of bounding box prediction \cite{pratt2020grounded} similar to value and value-all.
A predicted bounding box is correct if the overlap with the ground truth is $\ge$ 0.5.
}
{The metrics value, value-all, grnd value and grnd value-all are evaluated in three settings based on whether we are using ground truth verb, top-1 predicted verb, or top-5 predicted verbs.
For our model ablation on semantic role labeling and noun localization, we use the ground truth verb setting for measuring value, value-all, grnd value, and grnd value-all.
All experiments are performed on the dev set unless otherwise specified.}

\subsection{Implementation Details on imSitu}
We use the CLIP model with ViT-B32 image encoder to extract image features unless otherwise specified. 
The input to ClipSitu MLP is a concatenation of the CLIP embeddings of the image, verb, and role, each of 512 dimensions leading to 1536 dimensions. 
The input to ClipSitu TF is a sequence of tokens where each token is the concatenated image, verb, and role CLIP embeddings similar to ClipSitu MLP. 
To reduce the size of each token we project it to 512 dimensions using a linear layer.
We set the sequence length for ClipSitu TF to be 6 which refers to the maximum number of roles possible for a verb following \cite{cooray2020attention}.
Each verb has a varying number of roles and we mask the inputs when a verb has less than 6 roles.
The key and value for our patch-based cross-attention Transformer (ClipSitu XTF) is the unpooled image patch embeddings from CLIP image encoder .
For example, when using the ViT-B32 model as the CLIP image encoder, we end up with 50 image patch embeddings ($224/32 \times 224/32$ + 1 class) of 512 dimensions that are used as key and value. 
The query of ClipSitu XTF is a sequence of concatenated verb and role CLIP embeddings (1024 dimensions) tokens where each token is projected to 512 dimensions using a linear layer.
Similar to ClipSitu TF, the sequence length of the query in ClipSitu XTF is set to 6 and we mask the inputs when a verb has less than 6 roles.
Unless otherwise mentioned, we train all our models with a batch size of 64, a learning rate of 0.001, and  an ExponentialLR scheduler with Adamax optimizer, on a 24 GB Nvidia 3090.
In imSitu dataset \cite{yatskar2016situation}, we have 504 unique verbs, 190 unique roles, and 11538 unique nouns.
We extract CLIP text embeddings each verb, role, and noun separately.

\subsection{Analysis on imSitu with CLIP Image Encoders}
\textbf{Verb Prediction Comparison.} In \cref{tab:verbablation}, we compare the proposed ClipSitu Verb MLP model against zero-shot and linear probe performance of CLIP. 
We also compare against a CLIP finetuning model called weight-space ensembles (wise-ft) \cite{wortsman2022robust} that leverages both zero-shot and fine-tuned CLIP models to make verb predictions.
We compare ClipSitu Verb MLP and wise-ft using 4 CLIP image encoders - ViT-B32, ViT-B16, ViT-L14, and ViT-L14@336px.
The image clip embeddings for ViT-B32 and ViT-B16 are 512 dimensions and for ViT-L14, and ViT-L14@336px are 768 dimensions.
These four encoders represent different image patch sizes, different depths of image transformers, and different input image sizes.
We set the hidden layer dimension to 1024 in the ClipSitu Verb MLP.
The zero-shot performance of CLIP suggests that CLIP image features are beneficial for situation recognition tasks.
The best performance of ClipSitu Verb MLP is obtained with a single hidden layer and increasing the number of hidden layers does not improve performance.
Our best performing ClipSitu Verb MLP outperforms linear probe by 4.46\% on Top-1 and wise-ft by 3.2\% on Top-5 when using the same ViT-L14 image encoder.
ClipSitu Verb MLP performs better than wise-ft and linear probe for all image encoders. Therefore, verb prediction benefits from a well-designed MLP model such as our ClipSitu Verb MLP than finetuning the CLIP image encoder itself (wise-ft) or using regression (linear probe) for situational verb prediction.

\begin{table}[t]
\centering
\scriptsize
\resizebox{\linewidth}{!}{
\begin{tabular}{l|l|c|c|c} 
\hline
\begin{tabular}[c]{@{}l@{}}Image\\ Encoder\end{tabular} & Verb Model & \begin{tabular}[c]{@{}c@{}}Hidden\\ Layer\end{tabular} & Top-1 & Top-5 \\ 
\hline
\multirow{6}{*}{ViT-B32} & {zero-shot} & {-} & {29.2} & {65.2} \\ 
\hhline{~----}
 & {linear probe} & {-} & {44.6} & {78.4} \\ 
\hhline{~----}
 & wise-ft & - & 46.5 & 74.3 \\ 
\hhline{~----}
 & \multirow{3}{*}{ClipSitu Verb MLP} & 1 & 46.7 & 76.1 \\
 &  & 2 & 46.5 & 76.1 \\
 &  & 3 & 44.5 & 74.2 \\ 
\hline
\multirow{6}{*}{ViT-B16} & {zero-shot} & {-} & {31.9} & {67.89} \\ 
\hhline{~----}
 & {linear probe} & {-} & {49.3} & {78.8} \\ 
\hhline{~----}
 & wise-ft & - & 48.8 & 83.5 \\ 
\hhline{~----}
 & \multirow{3}{*}{ClipSitu Verb MLP} & 1 & 50.9 & 89.6 \\
 &  & 2 & 50.8 & 89.4 \\
 &  & 3 & 48.6 & 88.5 \\ 
\hline
\multirow{4}{*}{ViT-L14} & {zero-shot} & {-} & {38.2} & {79.3} \\ 
\hhline{~----}
 & {linear probe} & {-} & {52.4} & {87.7} \\ 
\hhline{~----}
 & wise-ft & - & 51.5 & 84.3 \\ 
\hhline{~----}
 & \multirow{3}{*}{ClipSitu Verb MLP} & 1 & 56.7 & 84.6 \\
 &  & 2 & 56.6 & 84.5 \\
 &  & 3 & 53.8 & 82.4 \\ 
\hline
\multirow{6}{*}{\begin{tabular}[c]{@{}l@{}}ViT-L14\\@336px\end{tabular}} & {zero-shot} & {-} & {39.7} & {79.2} \\ 
\hhline{~----}
 & {linear probe} & {-} & {53.4} & {81.4} \\ 
\hhline{~----}
 & wise-ft & - & 52.2 & 82.9 \\ 
\hhline{~----}
 & \multirow{3}{*}{ClipSitu Verb MLP} & 1 & \textbf{57.9} & \textbf{86.1} \\
 &  & 2 & 56.2 & 84.5 \\
 &  & 3 & 54.3 & 82.8 \\
\hline
\end{tabular}
}
\caption{Comparing performance of ClipSitu Verb MLP with zero-shot and finetuned CLIP (linear probe and wise-ft \cite{wortsman2022robust}).}
\label{tab:verbablation}
\end{table}
\begin{table}[t]
\centering
\scriptsize
\resizebox{\linewidth}{!}{
\begin{tabular}{l|l|cc|cc|cc}
\hline
\multirow{2}{*}{\begin{tabular}[c]{@{}l@{}}Image\\ Encoder\end{tabular}} & \multirow{2}{*}{\begin{tabular}[c]{@{}c@{}} Model\end{tabular}} & \multicolumn{2}{c|}{Top-1 } & \multicolumn{2}{c|}{Top-5} & \multicolumn{2}{c}{Ground truth} \\ \hhline{~~------} 
 & & value & v-all & value & v-all & value & v-all \\ \hline
{\multirow{3}{*}{ViT-B32}} & {MLP} & 45.6 & 27.1 &  66.3 & 37.5 & 76.9 & 43.2 \\
{} & {TF} & 45.7 & 27.3 & 66.3 & 38.0 & 76.8 & 43.0 \\
{} & {XTF} & 44.5 & 25.9 & 64.9 & 35.6 & 75.2 & 40.8 \\ \hline
{\multirow{3}{*}{ViT-B16}} & {MLP} & 46.3 & 28.2 & 67.3 & 39.4 & 77.9 & 44.8 \\
{} & {TF} & 46.4 & 28.6 & 67.4 & 39.8 & 77.2 & 43.8 \\
{} & {XTF} & 45.7 & 27.4 & 66.1 & 37.4 & 75.4 & 40.6 \\ \hline
{\multirow{3}{*}{ViT-L14}} & { MLP} & 46.5 & 28.4 & 67.6 & 39.7 & 77.6 & 43.9 \\
{} & {TF} & 46.9 & 29.6 & 68.2 & 41.2 & 78.0 & 45.2 \\
{} & {XTF} &  46.9 & 29.5 & 68.1 & 40.6 & 77.8 & 44.5 \\ \hline
\multirow{3}{*}{\begin{tabular}{@{}l@{}}ViT-L14\\@336px\end{tabular}} & {MLP} & 46.7  & 29.1 & 67.9 & 40.5 & 77.9 & 44.9 \\
{} & {TF} & 47.0  & 29.7 & 68.3 & 41.4 & 78.3 & 45.8 \\
{} &  XTF & \textbf{47.2} & \textbf{30.1} & \textbf{68.4} & \textbf{41.7} & \textbf{78.5} & \textbf{45.8} \\ \hline
\end{tabular} }
    \caption{Comparison of CLIP Image Encoders on noun prediction task using top-1 and top-5 predicted verb from  the best-performing Verb MLP model obtain from ~\cref{tab:verbablation}. All models' performance improves by increasing the number of patch tokens either by reducing patch size (32$\rightarrow$16$\rightarrow$14) or increasing image size (224$\rightarrow$336). v-all stands for value all.
    } 
    \label{tab:clipablation}
\end{table}

\textbf{Noun Prediction Comparison.} Next, we study the effect of using different CLIP image encoders for noun prediction with ClipSitu MLP, TF and XTF.
Again as verb prediction, we use four image encoders for a thorough comparison.
Please note that the number of image patch embeddings used as key and value changes based on patch size and image size for ClipSitu XTF.
We obtain 197 image patch embeddings ($224/16 \times 224/16$ + 1 class token) for ViT-B16, 257 image patch embeddings for ViT-L14 ($224/14 \times 224/14$ + 1 class token), and 
577 image patch embeddings for ViT-L14@336px ($336/14 \times 336/14$ + 1 class token).
For ViT-L14, and ViT-L14@336px image encoders, we obtain 768-dimensional embeddings which we project using a linear layer to 512 dimensions.
We choose the best hyper-parameters for ClipSitu MLP, TF, and XTF based on the ablations presented in \cref{sec:archabl}.

In \cref{tab:clipablation}, we observe that the value and value-all using ground truth verbs steadily improve for all three models as the number of patches increase from 50 (ViT-B32) to 577 (ViT-L14@336px) or the image size increases from 224$\times$224 to 336$\times$336.
For smaller backbones such as ViT-B32 and ViT-B16, the best performance is obtained by ClipSitu MLP while ClipSitu XTF shows the most improvement when using a larger backbone (ViT-L14) and the largest image size (ViT-L14@336px).
ClipSitu XTF is able to extract more relevant information when attending to more image patch tokens to produce better predictions.
To compare noun prediction using top-1 and top-5 predicted verbs, we use the best ClipSitu Verb MLP (ViT-L14@336px) from \cref{tab:verbablation}.
Even when using Top-1 and Top-5 predicted verbs, we observe a similar trend as the ground truth verb. 
ClipSitu XTF again shows the most improvement in value and value-all to obtain the best performance among the three models across ground truth, Top-1 and Top-5 predicted verbs.

\subsection{VidSitu Evaluation Details}\label{sec:vidsitu_eval_det}
Semantic role labeling in VidSitu is evaluated using generation metrics such CIDEr, Rouge-L, and Lea.
VidSitu comprises of 23,626 training and 1326 validation videos.
Following \cite{sadhu2021visual}, we generate the nouns 4 roles at most for every verb depicted by placeholders Arg0, Arg1, Arg2, ArgScn (Scene) that vary for every verb.
The generation metrics used for comparison are CIDEr \cite{vedantam2015cider} macro-averaged for verb (C-Vb), and macro-averaged CIDEr (C-Arg), Rouge \cite{lin2004rouge} (R-L) and Lea \cite{moosavi2016coreference} for noun.

\subsection{Implementation Details on VidSitu} \label{sec:vidsitu_impl}
For video embedding, we use X-CLIP \cite{ma2022x} which is a minimal extension to CLIP for videos that captures similarities between video and text at both coarse-grained (video-sentence level) and fine-grained (frame-word level).
One of the main distinctions in transformer decoder (TxD) comprises of 3 layers with 8 attention heads inspired by existing works \cite{sadhu2021visual, xiao2022hierarchical}.
The encoder outputs from MLP, TF, and XTF are used as encoder attention to start the decoding. 
We use greedy decoding with temperature 1.0 instead of beam search as it provides better performance.

\subsection{Analysis on VidSitu with different X-CLIP Encoders} \label{sec:vidsitu_abl}
\begin{table*}
\centering
\scriptsize
\begin{tabular}{l|c|c|c|c|c|c} 
\hline
Method & Feature & CIDEr & C-Vb & C-Arg & R-L & Lea \\ 
\hline
ClipSituMLP+TxD & \multirow{3}{*}{X-CLIP (ViT-B32)} & 60.53 & \textbf{71.21} & 52.80 & 43.29 & 37.30 \\ 
ClipSituTF+TxD &  & 58.97 & 67.75 & 49.81 & 42.25 & 50.19 \\ 
ClipSituXTF+TxD & & 57.75 & 66.17 & 47.04 & 41.71 & 49.11 \\ 
\hline
ClipSituMLP+TxD & \multirow{3}{*}{X-CLIP (ViT-L14)} & \textbf{61.93} & 70.14 & \textbf{56.30} & \textbf{43.77} & 37.77 \\ 
ClipSituTF+TxD &  & 59.03 & 68.07 & 53.51 & 39.93 & 45.29 \\ 
ClipSituXTF+TxD &  & 49.65 & 58.35 & 44.91 & 42.40 & \textbf{50.91} \\
\hline
\end{tabular}
\caption{Comparison of ClipSitu MLP, TF and XTF models on VidSitu with different X-CLIP features. ClipSitu MLP+TxD performs the best on 4 out of 5 generation metrics.}
\label{tab:vidsitu_ablation}
\end{table*}
We compare the performance of ClipSitu MLP, TF and XTF with the decoder(TxD) in \cref{tab:vidsitu_ablation}. 
We use two X-CLIP features -- ViT-B32 and ViT-L14 for thorough testing of our models.
ClipSituMLP+TxD performs the best among all the models on 4 out of 5 generation metrics using either of the X-CLIP features.
We attribute this to ClipSituMLP making the least changes to the input X-CLIP features compared to ClipSituTF and XTF.
This property also allows ClipSituMLP+TxD to perform better with a more descriptive feature such as ViT-L14 compared to ViT-B32.

\subsection{Ablations on imSitu}\label{sec:archabl}
\begin{table*}[t]
\resizebox{\linewidth}{!}{
\begin{tabular}{ll|cccc|cccc|cccc|cccc}
\hline
Heads &  & \multicolumn{4}{c|}{1} & \multicolumn{4}{c|}{2} & \multicolumn{4}{c|}{4} & \multicolumn{4}{c}{8}\\ \hline
Layers &  & 1 & 2 & 4 & 6 & 1 & 2 & 4 & 6 & 1 & 2 & 4 & 6 & 1 & 2 & 4 & 6\\ \hline
\multirow{2}{*}{ClipSitu TF} & value & 75.73 & 75.78 & \textbf{76.87} & 24.68 & 75.80 & 75.95 & 75.97 & 18.28 & 75.71 & 76.77 & 75.87 & 05.20 & 75.74 & 75.93 & 76.07 & 75.94\\
 & value-all & 41.40 & 41.52 & 41.84 & 00.21 & 41.64 & 41.60 & 41.83 & 00.21 & 41.43 & 42.10 & 41.75 & 00.00 & 41.35 & 41.58 & \textbf{42.97} & 41.72\\ 
 \hline
\multirow{2}{*}{ClipSitu XTF} & value & 72.70 & 74.33 &  \textbf{75.27} & 74.35 & 53.11 & 53.17 & 53.11 & 53.16 & 53.18 & 53.51 & 53.45 & 53.49 & 53.13 & 53.44 & 53.38 & 53.54\\
 & value-all & 36.61 & 39.11 & \textbf{40.79} & 39.06 & 16.58 & 16.64 & 16.38 & 16.46 & 16.50 & 16.77 & 16.90 & 16.97 & 16.42 & 16.89 & 16.85 & 17.02 \\
\hline
\end{tabular}}
\caption{Ablation on Transformer hyperparameters. 1 head with 4 layers is sufficient to obtain best value and value-all performance for ClipSitu XTF. For TF, 1 head and 4 layers produces best value whereas 8 heads and 4 layers produces best value-all performance.}
\label{tab:tfablation}
\end{table*}

\begin{figure}
    \begin{tabular}{l}
    \includegraphics[width=0.48\linewidth]{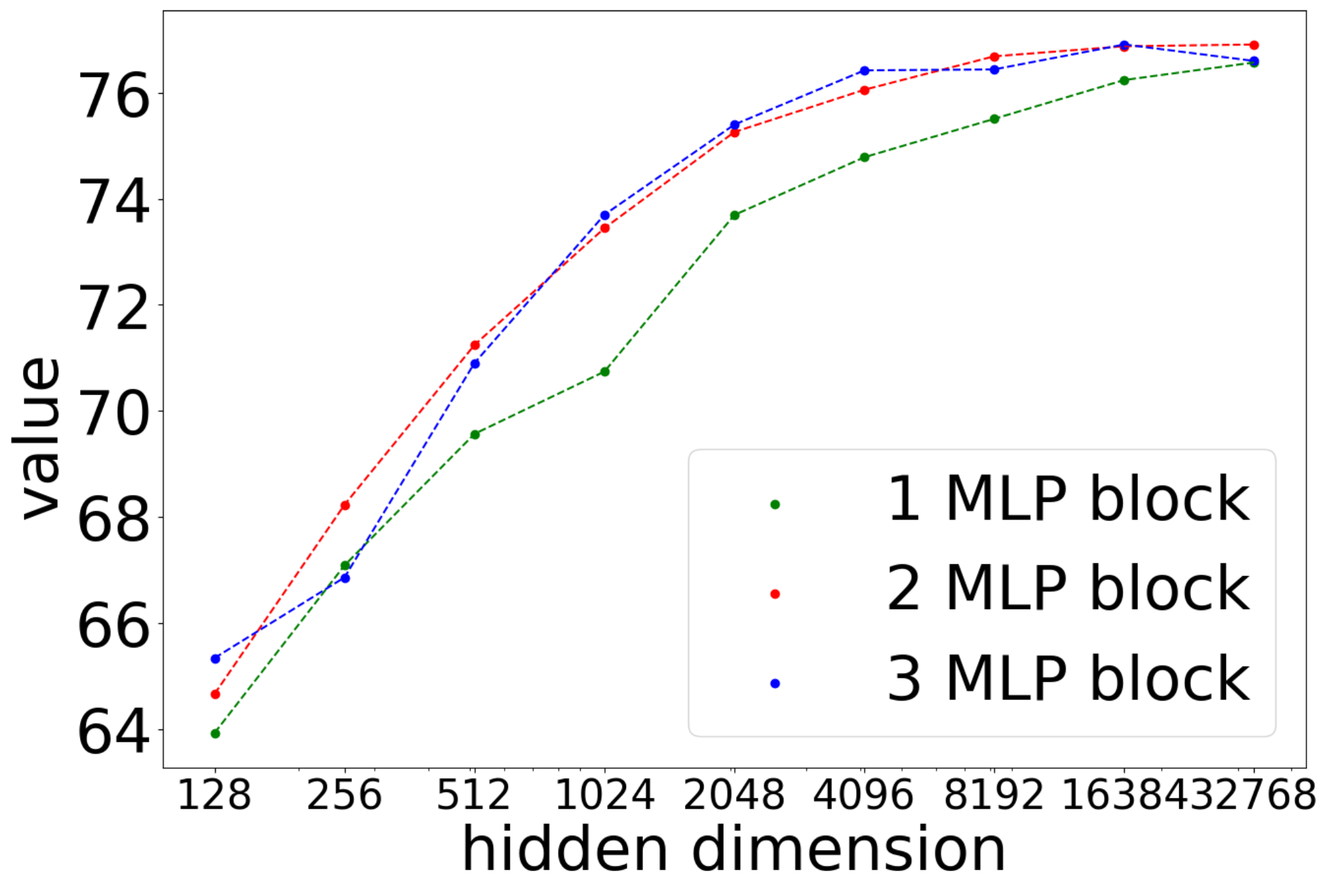}  \includegraphics[width=0.48\linewidth]{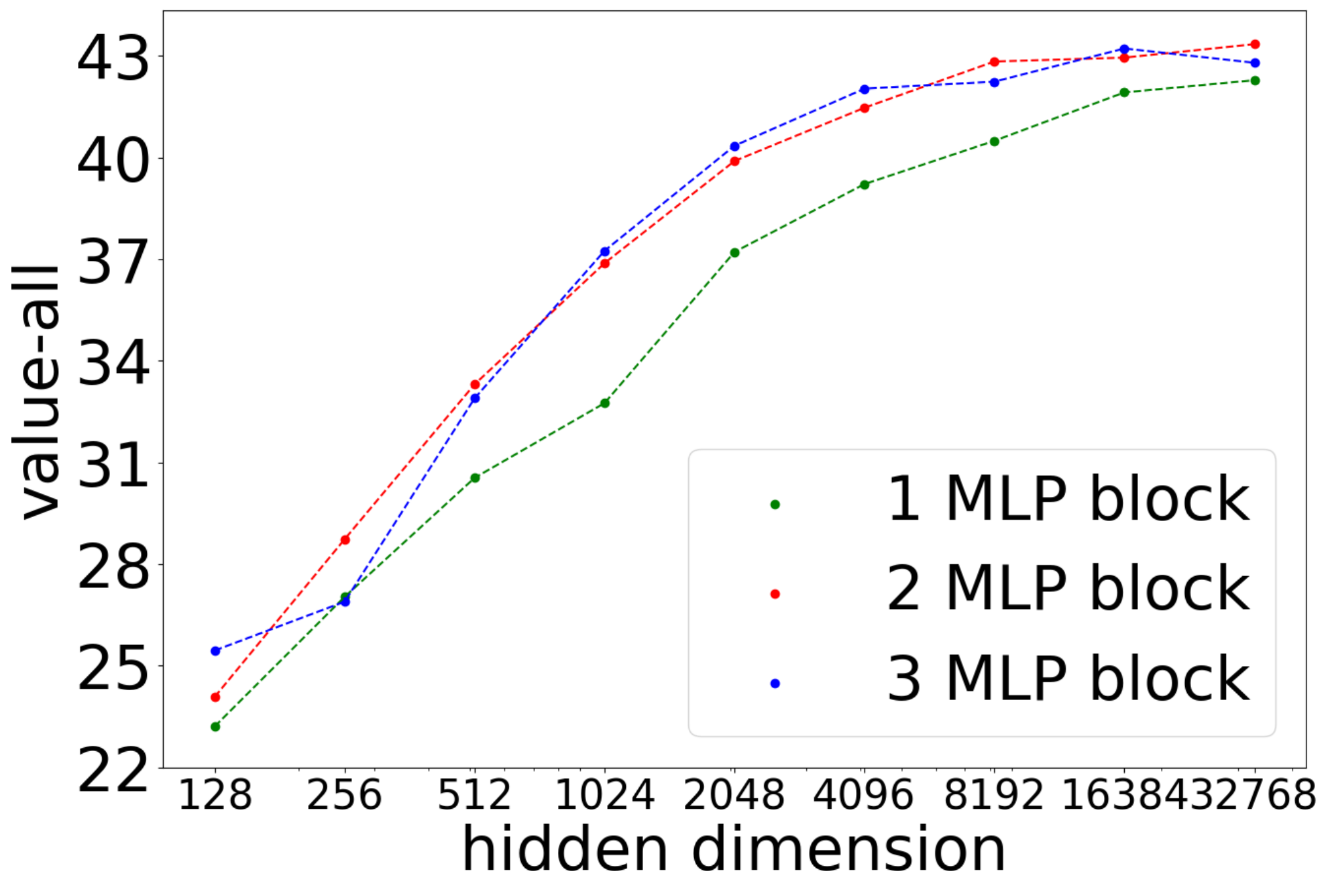} \\
    \end{tabular}
    \caption{Effect of the number of MLP blocks and hidden dimensions on value and value-all. We train with very large hidden dimensions such as 8192, 16384, and 32768 to obtain state-of-the-art value and value-all results. }
    \label{fig:mlpablation}
\end{figure}

\textbf{ClipSitu MLP ablation} In \cref{fig:mlpablation}, we explore combinations of MLP blocks and the dimension of the hidden (linear) layer in each block to obtain the best MLP network for semantic role labeling. 
We train ClipSitu MLP with small to very large hidden layer dimensions i.e. 128$\rightarrow$16384 which results in a steady improvement in both value and value-all.
Increasing the number of MLP blocks also improves the performance as a model that suggests we need a model with more parameters to predict the large number of unique nouns -- 11538.
However, we reach saturation at hidden layer dimension of 32768 with 3 MLP blocks as no further improvement in value and value-all is observed.
Our best ClipSitu MLP for semantic role labeling obtains 76.91 for value and 43.22 for value-all with 3 MLP blocks with each block having 16,384 hidden dimensions which beats the state-of-the-art CoFormer \cite{cho2022collaborative}.
The main reason our ClipSitu MLP performs well on semantic role labeling is our modern MLP block design that contains large hidden dimensions along with LayerNorm which have not been explored in existing MLP-based CLIP finetuning approaches such as wise-ft \cite{wortsman2022robust}.
We also compare the performance of ClipSitu MLP with our proposed minimum annotator cross-entropy loss ($\mathcal{L}_{MAXE}$) versus applying cross-entropy using the noun labels of each annotator separately.
We find that $\mathcal{L}_{MAXE}$ produces better value and value-all performance (76.91 and 43.22) compared to cross-entropy (76.57 and 42.88).

\begin{table}
\scriptsize
\centering
\resizebox{\linewidth}{!}{
\begin{tabular}{l|cc|cc|cc}
\hline
\multirow{2}{*}{\begin{tabular}[c]{@{}c@{}} {Model}\end{tabular}} & \multicolumn{2}{c|}{{Top-1} } & \multicolumn{2}{c|}{{Top-5}} & \multicolumn{2}{c}{{Ground truth}} \\ \hhline{~|------}
  & \begin{tabular}[c]{@{}l@{}}{grnd}\\ {value}\end{tabular} & \begin{tabular}[c]{@{}l@{}}{grnd}\\ {v-all}\end{tabular} & \begin{tabular}[c]{@{}l@{}}{grnd}\\ {value}\end{tabular}& \begin{tabular}[c]{@{}l@{}}{grnd}\\ {v-all}\end{tabular} & \begin{tabular}[c]{@{}l@{}}{grnd}\\ {value}\end{tabular} & \begin{tabular}[c]{@{}l@{}}{grnd}\\ {v-all}\end{tabular} \\ \hline
verb-role emb & 18.91 & 05.34 & 22.45 & 17.75 & 27.81 & 20.07 \\ \hline
  {XAtt. L1} & {39.30} & {10.54} & {46.46} & {19.70} & {53.87} & {30.22} \\
  {XAtt. L4} & {{33.30} } & { {09.34}} & {{42.55}} & {{17.53}} & 
{{51.32}} & {{31.32}}\\
  \begin{tabular}[c]{@{}l@{}}{XAtt. L1  + L4} \end{tabular} & {{36.56} } & { {09.88}} & {{44.23}} & {{11.43}} & 
{{54.56}} & {{34.71}}\\ \hline
  \begin{tabular}[c]{@{}l@{}} {verb-role emb }\\ {+ XAtt L1}\end{tabular} & {\textbf{41.30} } & { \textbf{13.92}} & {\textbf{49.23}} & {\textbf{23.45}} & 
{\textbf{55.36}} & {\textbf{32.37}}\\ \hline
 \end{tabular}
 }

\caption{{Comparing verb-role embeddings with different ClipSitu XTF inputs for noun localization. XAtt. -- cross-attention scores. L1 -- first XTF layer and L4 -- last XTF layer of best performing model (1 head, 4 layers, ViT-L14@336px). Cross-attention scores vastly improves the performance of noun localization. We use the best Verb MLP from \cref{tab:verbablation}. v-all stands for value-all.}}
\label{tab:grndcompare}

\end{table}


\textbf{ClipSitu TF and XTF ablation} In \cref{tab:tfablation}, we explore the number of heads and layers needed in ClipSitu TF and XTF to obtain the best performance in semantic role labeling. 
We find that a single head with 4 transformer layers performs the best in terms of value for both ClipSitu TF and XTF while for value-all, an 8-head 4-layer ClipSitu TF performs the best and we use this for subsequent evaluation.
For both ClipSitu TF and XTF, increasing the number of layers beyond 4 does not yield any improvement in value or value-all when using less number of heads (1,2,4).
Similarly, for ClipSitu XTF, increasing the number of heads and layers leads to progressively deteriorating performance.
Both of these performance drops can be attributed to the fact that we have insufficient samples for training larger transformer networks \cite{radford2021learning}.

\textbf{Comparison with other multimodal embeddings} We compare the performance of CLIP with another multimodal embedding model -- ALIGN \cite{jia2021scaling} which is trained to align images with their captions \cref{tab:compare_embed}. 
We also compare CLIP text embedding with other text embedding methods such as Glove \cite{pennington2014glove} and BERT \cite{kenton2019bert} to obtain verb and role embeddings.
ALIGN does not perform as well as CLIP on semantic role prediction perhaps because it does not capture objects well in the images as it trained mainly on alt-text data that may not describe all the objects in the image.
Glove and BERT verb and role embeddings with CLIP image embeddings perform well on both verb prediction and semantic role prediction.
Therefore, ClipSitu XTF is flexible enough to incorporate any text embedding model and can improve with better text embedding models.

\begin{table*}
\centering
\scriptsize
\resizebox{\linewidth}{!}{
\begin{tabular}{ll|ccc|ccc|cc}
\hline
\multirow{2}{*}{Verb \& Role embedding} & \multirow{2}{*}{Image emb.} & \multicolumn{3}{c|}{Top-1 predicted verb} & \multicolumn{3}{c|}{Top-5 predicted verb} & \multicolumn{2}{c}{Ground truth verb} \\ \hhline{~~--------}
& & verb & value & v-all & verb & value & value-all & value & value-all \\ \hline
ALIGN \cite{jia2021scaling} & ALIGN & 52.27 & 31.17 & 12.17 &	80.56 &	45.26 &	15.75 &	52.95 &	16.86 \\
Glove \cite{pennington2014glove} & CLIP & 55.03 &	42.40 &	24.27 &	82.96 &	62.60 &	33.94 &	73.44 &	37.89 \\ 
BERT \cite{kenton2019bert} & CLIP & 55.03 & 42.10 & 23.47 & 82.96 & 62.15 & 33.09 & 73.03 & 37.24 \\
CLIP & CLIP & \textbf{58.19} & \textbf{47.23} & \textbf{29.73} & \textbf{85.69} & \textbf{68.42} & \textbf{41.42} & \textbf{78.52} & \textbf{45.31} \\ 
\hline
\end{tabular}
}
\caption{Comparing different text and image embedding models on the XTF model. Glove and BERT verb and role embedding perform well with CLIP image embeddings.}
\label{tab:compare_embed}
\end{table*}

\textbf{Complexity} We compare the number of parameters, computation, and inference time for ClipSitu MLP, TF, and XTF using the ViT-L14-336 image encoder and CoFormer \cite{cho2022collaborative} in \cref{tab:params}.
We find that ClipSitu TF is the most efficient in terms of parameters, computation, and inference time closely followed by ClipSitu XTF at half the parameters of CoFormer and 9\% of ClipSitu MLP.
Even adding the lightweight ClipSitu Verb MLP to ClipSitu XTF for combined verb and noun prediction leads to a very efficient but effective model.
Therefore, we conclude that ClipSitu XTF not only performs the best at semantic role labeling but is also efficient in terms of parameters, computation, and inference time compared to ClipSitu MLP and CoFormer \cite{cho2022collaborative}.
\begin{table}[]
    \centering
    \scriptsize
    \resizebox{\linewidth}{!}{
    \begin{tabular}{l|c|c|c}
    \hline
    Model & \# Parameters & {GFlops} & {Inference Time(ms)}\\
    \hline
         {CoFormer \cite{cho2022collaborative}} & {93.0M} & {1496.67} & 
         {30.62} \\
         ClipSitu Verb MLP & {1.3M}	& {0.17}	& {0.08} \\ 
         ClipSitu MLP &  580.2M & {443.18} & {32.33} \\
         ClipSitu TF & \textbf{20.2M} & {\textbf{8.65}}  &  {\textbf{1.55}}\\
         ClipSitu XTF & 45.3M & {116.01} & {11.17}\\
         \hline
    \end{tabular}
    }
    \caption{Comparison of parameters, flop count and inference time for CoFormer \cite{cho2022collaborative}, Verb MLP, ClipSitu MLP, TF and XTF models.}
    \label{tab:params}
\end{table}
\begin{figure}
    \centering
    \includegraphics[width=\linewidth]{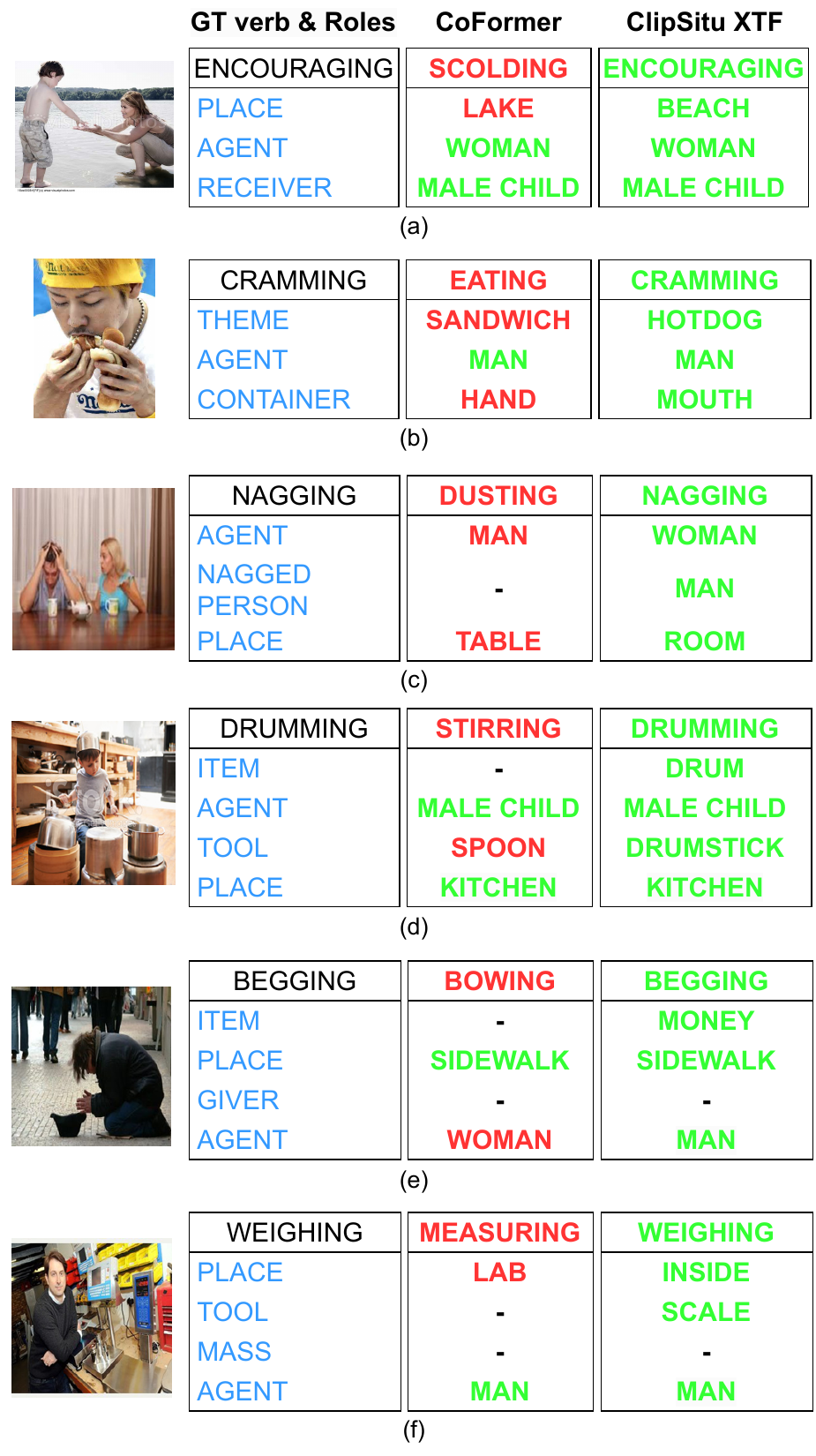}
    \caption{Qualitative comparison of ClipSitu XTF and CoFormer \cite{cho2022collaborative} predictions. \textcolor{green}{green} refers to correct prediction while \textcolor{red}{red} refers to incorrect prediction. '-' refers to predicting blank (a noun class) for this role.}
    \label{fig:qual_noun}
\end{figure}

\begin{figure}
    \centering
    \includegraphics[width=0.7\linewidth]{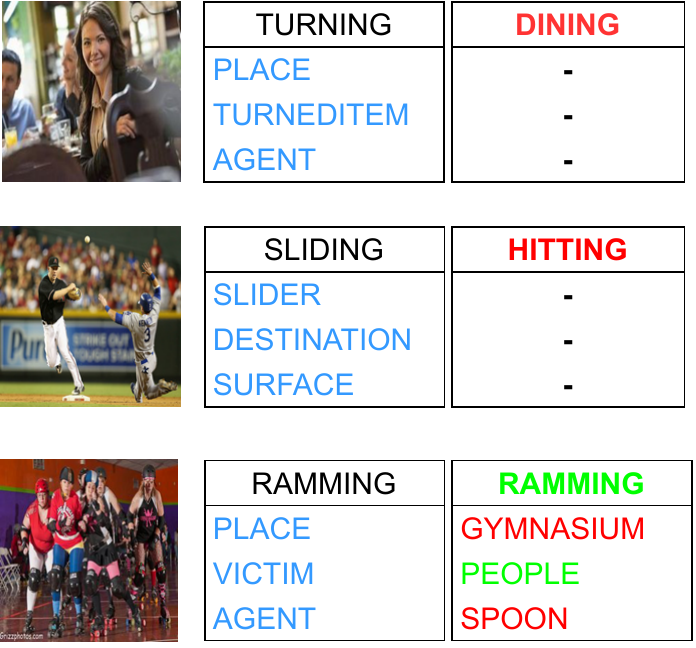}
    \caption{Examples where ClipSitu XTF predicts the verb or noun incorrectly.}
    \label{fig:fail}
\end{figure}
\textbf{Qualitative Comparison of XTF.}
In \cref{fig:qual_noun}, we compare the qualitative results of ClipSitu XTF with CoFormer. 
ClipSitu XTF is able to correctly predicts verbs such as cramming (\cref{fig:qual_noun}(b)) while CoFormer focuses on the action of eating and hence incorrectly predicts the verb which also makes its noun predictions for the container and theme incorrect.
CoFormer predicts the place as table and predicts the verb as dusting (\cref{fig:qual_noun}(c)) instead of focusing on the action of nagging. 
Finally, we see in \cref{fig:qual_noun}(d) that CoFormer is confused by the visual context of kitchen as it predicts stirring instead of identifying the action which is drumming. 
On the other hand, ClipSitu XTF correctly predicts drumming and the tool as drumsticks while still predicting place as kitchen. 
In \cref{fig:fail}, we show some qualitative examples where ClipSitu Verb MLP predicts the verb incorrectly or ClipSitu XTF predicts the noun incorrectly.

\begin{figure}
    \centering
    \scriptsize
    \begin{tabular}{cc}
    \includegraphics[width=0.45\linewidth]{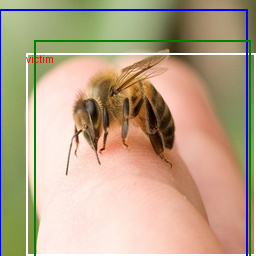} &
    \includegraphics[width=0.45\linewidth]{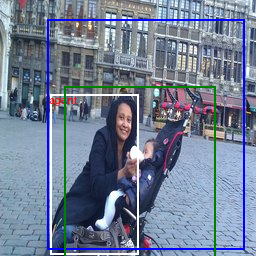} \\
    (a) verb: stinging & (b) verb: feeding\\
    role: agent & role: agent\\
    \includegraphics[width=0.45\linewidth]{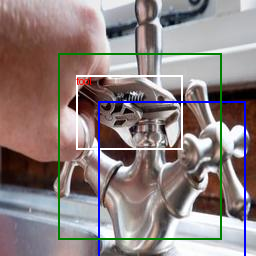} &
    \includegraphics[width=0.45\linewidth]{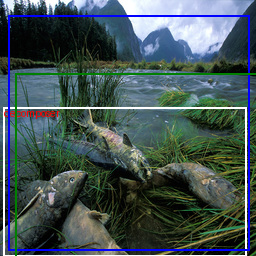} \\
    (c) verb: fixing & (d) verb: decomposing \\
    role: tool & role: decomposer\\
    \includegraphics[width=0.45\linewidth]{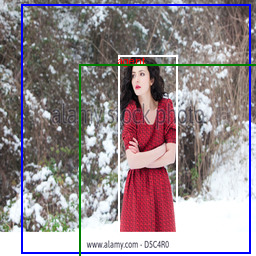} &
    \includegraphics[width=0.45\linewidth]{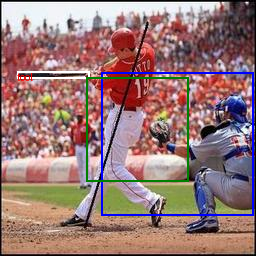} \\
    (e) verb: shivering & (f) verb: hitting\\
    role: agent & role: tool\\
    \end{tabular}
    \caption{Examples of noun localization when using verb-role embeddings in \textcolor{blue}{blue} vs. cross-attention scores in \textcolor{green}{green}. In (a) to (e), cross-attention scores localize the noun better than verb-role embeddings with the bounding boxes closer and tighter around the noun (ground-truth bounding box in white). When the noun is very small compared to the image (baseball bat in (f)) it is challenging for both methods to localize the noun.
    Cross-attention scores from Layer 1 (XAtt. L1 in \cref{tab:grndcompare}) }
    \label{fig:qual_grnd}
\end{figure}

\textbf{Noun Localization ablation.} In \cref{tab:grndcompare}, we present noun localization results of ClipSitu XTF.
First, we show that cross-attention scores vastly outperform verb-role embeddings. 
Second, cross-attention scores taken from the first (XAtt. L1) performs better than the fourth layer (XAtt. L4) indicating that cross-attention score is more useful for localization the less transformation it has undergone i.e. closer to the original image.
Cross-attention scores are obtained from the patch-based CLIP image features as key and value and verb-role embeddings as query.
Therefore, the cross-attention score represents the similarity of each patch to the role. 
The similarity information helps accurately predict the patches corresponding to each noun.
We also concatenate verb and role embeddings to the cross-attention score (XAtt. L1 + verb-role) to provide more context about the role which further improves localization performance.
We visualize these quantitative noun localization results using  qualitative examples \cref{fig:qual_grnd}(a)-(e) of noun localization.
We visualize a random role per image where the noun is predicted correctly by ClipSitu XTF. 
We compare the bounding box predicted for that noun using verb-role embeddings and cross-attention scores (XAtt. L1).
Cross-attention score produces bounding boxes closer to the ground-truth than verb-role embedding.
We also show a challenging in \cref{fig:qual_grnd}(f) image where the noun occupies a very small region in the image where both cross-attention and verb-role embeddings are not able to localize it accurately.




\revise{\section{Comparison with state-of-the-art}}

\revise{\subsection{Image Situation Recognition}\label{sec:imsitu_sota}} 
\begin{table*}
\centering
\scriptsize
\resizebox{\linewidth}{!}{
\begin{tabular}{ll|ccccc|ccccc|cccc}
\hline
 &  & \multicolumn{5}{c|}{Top-1 predicted verb} & \multicolumn{5}{c|}{Top-5 predicted verb} & \multicolumn{4}{c}{Ground truth verb} \\ \hhline{~~--------------}
 
\multirow{-2}{*}{Set} & \multirow{-2}{*}{Method} & verb & value & v-all & { {\begin{tabular}[c]{@{}l@{}}grnd\\ value\end{tabular}}}& { {\begin{tabular}[c]{@{}l@{}}grnd\\ v-all\end{tabular}}} & verb & value & value-all & { {\begin{tabular}[c]{@{}l@{}}grnd\\ value\end{tabular}}}  & { {\begin{tabular}[c]{@{}l@{}}grnd\\ value-all\end{tabular}}} & value & value-all & { {\begin{tabular}[c]{@{}l@{}}grnd\\ value\end{tabular}}} & { {\begin{tabular}[c]{@{}l@{}}grnd\\ value-all\end{tabular}}} \\ \hline

 & CRF \cite{yatskar2016situation} & {32.25} & {24.56} & 14.28 & { -} & { -} & {58.64} & {42.68} & 22.75 & { -} & { -} & {65.90} & 29.50 & { -} & { -} \\
 & RNN w/ Fus. \cite{mallya2017recurrent} & {36.11} & {27.74} & 16.60 & { -} & { -} & {63.11} & {47.09} & 26.48 & { -} & { -} & {70.48} & 35.56 & { -} & { -} \\
 & GraphNet \cite{li2017situation} & {36.93} & {27.52} & 19.15 & { -} & { -} & {61.80} & {45.23} & 29.98 & { -} & { -} & {68.89} & 41.07 & { -} & { -} \\
 & CAQ RE-VGG \cite{cooray2020attention} & {37.96} & {30.15} & 18.58 & { -} & { -} & {64.99} & {50.30} & 29.17 & { -} & { -} & {73.62} & 38.71 & { -} & { -} \\
 & Ker. GrphNet \cite{suhail2019mixture} & {43.21} & {35.18} & 19.46 & { -} & { -} & {68.55} & {56.32} & 30.56 & { -} & { -} & {73.14} & 41.68 & { -} & { -} \\
 & JSL \cite{pratt2020grounded} & {39.60} & {31.18} & 18.85 & { 25.03} & { 10.16} & {67.71} & {52.06} & 29.73 & { 41.25} & { 15.07} & {73.53} & 38.32 & { 57.50} & { 19.29} \\
 & GSRTR \cite{cho2021gsrtr} & {41.06} & {32.52} & 19.63 & { 26.04} & { 10.44} & {69.46} & {53.69} & 30.66 & { 42.61} & { 15.98} & {74.27} & 39.24 & { 58.33} & { 20.19} \\
 & SituFormer \cite{wei2022rethinking} & 44.32 & 35.35 & 22.10 & { 29.17} & { 13.33} & 71.01 & 55.85 & 33.38 & { 45.78} & { 19.77} & 76.08 & 42.15 & { \textbf{61.82}} & { 24.65} \\
 & CoFormer \cite{cho2022collaborative} & {44.41} & {35.87} & 22.47 & { 29.37} & { 12.94} & {72.98} & {57.58} & 34.09 & { 46.70} & { 19.06} & {76.17} & 42.11 & { 61.15} & { 23.09} \\ 
 \hhline{~---------------}
\multirow{-12}{*}{dev} & ClipSitu XTF & \multirow{1}{*}{\textbf{58.19}} & \textbf{47.23} & \textbf{29.73} & {\textbf{41.30} } & { \textbf{13.92}} & \multirow{1}{*}{\textbf{85.69}} & \textbf{68.42} & \textbf{41.42} & {\textbf{49.23}} & {\textbf{23.45}} & \textbf{78.52} & \textbf{45.31} &  
{55.36} & {\textbf{32.37}} \\ \hline

 & CRF \cite{yatskar2016situation} & {32.34} & {24.64} & 14.19 & { -} & { -} & {58.88} & {42.76} & 22.55 & { -} & { -} & {65.66} & 28.96 & { -} & { -} \\
 & RNN w/ Fus \cite{mallya2017recurrent} & {35.90} & {27.45} & 16.36 & { -} & { -} & {63.08} & {46.88} & 26.06 & { -} & { -} & {70.27} & 35.25 & { -} & { -} \\
 & GraphNet \cite{li2017situation} & {36.72} & {27.52} & 19.25 & { -} & { -} & {61.90} & {45.39} & 29.96 & { -} & { -} & {69.16} & 41.36 & { -} & { -} \\
 & CAQ RE-VGG \cite{cooray2020attention} & {38.19} & {30.23} & 18.47 & { -} & { -} & {65.05} & {50.21} & 28.93 & { -} & { -} & {73.41} & 38.52 & { -} & { -} \\
 & Ker. GrphNet \cite{suhail2019mixture} & {43.27} & {35.41} & 19.38 & { -} & { -} & {68.72} & {55.62} & 30.29 & { -} & { -} & {72.92} & 42.35 & { -} & { -} \\
 & JSL \cite{pratt2020grounded} & {39.94} & {31.44} & 18.87 & { {24.86}} & { {09.66}} & {67.60} & {51.88} & 29.39 & { {40.6}} & { {14.72}} & {73.21} & 37.82 & { {56.57}} & { {18.45}} \\
 & GSRTR \cite{cho2021gsrtr} & {40.63} & {32.15} & 19.28 & { {25.49}} & { {10.10}} & {69.81} & {54.13} & 31.01 & { {42.5}} & { {15.88}} & {74.11} & 39.00 & { {57.45}} & { {19.67}} \\
 & SituFormer \cite{wei2022rethinking} & 44.20 & 35.24 & 21.86 & { {29.22}} & { {13.41}} & 71.21 & 55.75 & 33.27 & { {46.00}} & { {20.10}} & 75.85 & 42.13 & { \textbf{61.89}} & { {24.89}} \\
 & CoFormer \cite{cho2022collaborative} & {44.66} & {35.98} & 22.22 & { {29.05}} & { {12.21}} & {73.31} & {57.76} & 33.98 & { {46.25}} & { {18.37}} & {75.95} & 41.87 & { {60.11}} & { {22.12}} \\
 & CLIP-Event \cite{li2022clip} & 45.60 & 33.10 & 20.10 & { 21.60} & {10.60} & - & - & - & {-} & { -} & - & - & { -} & { -} \\ 
 
 \hhline{~---------------}
 \multirow{-13}{*}{test} & ClipSitu XTF & \multirow{1}{*}{\textbf{58.09}} & \textbf{47.21} & \textbf{29.61} & {\textbf{40.01}} & {\textbf{15.03}} & 
 \multirow{1}{*}{\textbf{85.69}} & \textbf{68.42} & 
 \textbf{41.42} & {\textbf{49.78}} & {\textbf{25.22}} & \textbf{78.52} & \textbf{45.31} & {54.36} & {\textbf{33.20}} \\ \hline
\end{tabular}
}
\caption{Comparison with state-of-the-art on {Grounded Situation Recognition}. Robustness of ClipSitu MLP, TF, and XTF is demonstrated by the massive improvement for value and value-all with Top-1 and Top-5 predicted verbs over SOTA. 
}
\label{tab:sotatable}

\end{table*}

\revise{In \cref{tab:sotatable}, we compare the performance of proposed approaches with state-of-the-art approaches on both grounded and ungrounded situation recognition. 
Approaches that do not perform grounding and are used for comparison with our approach are as follows: (a) Conditional Random Field (CRF) \cite{yatskar2016situation} trained separately for verb prediction and noun prediction, (b) Recurrent Neural Network with fusion (RNN w/ Fus.) \cite{yatskar2017commonly} that uses CNN for verb prediction and an RNN for sequential prediction of nouns given the verb and image, (c) Graph Attention Network (GraphNet) \cite{li2017situation} containing interconnected verb and role nodes used for predicting the verb and nouns, respectively, (d) Context aware Reasoning 
(CAQ RE-VGG) \cite{cooray2020attention} utilizes a top-down attention mechanism to query the role from region features of an image image and the predicted verb from a CNN, (e) Mixture-Kernel Graph Attention Network (Ker. GraphNet) \cite{suhail2019mixture} extend GraphNet \cite{li2017situation} with context-aware interactions between role pairs across verbs through a set of shared basis kernel matrices, (f)  CLIP-Event \cite{li2022clip} uses CLIP for image encoding and GPT3 for decoding the entire situation description as a sentence containing the verb and the nouns.
The next set of approaches also perform noun grounding: (g) Joint Situation Localization (JSL) \cite{pratt2020grounded} jointly predicts verb, nouns and noun locations using a ResNet-LSTM network, (h) GSRTR \cite{cho2021gsrtr} uses a vision transformer encoder for verb prediction and a transformer decoder for noun prediction and localization, (i) SituFormer \cite{wei2022rethinking} extends GSRTR by re-ranking predicted nouns for every role using other examples of the same verb, (j) Collaborative Transformer (CoFormer) \cite{cho2022collaborative} is similar to SituFormer but decomposes noun prediction and localization by predicting nouns first and then their locations based on the verb and predicted nouns based on the image.
}

We use ViT-L14@336px image encoder for both ClipSitu Verb MLP and ClipSitu XTF.
ClipSitu Verb MLP outperforms SOTA method CoFormer on Top-1 and Top-5 verb prediction by a large margin of 12.6\% and 12.4\%, respectively, on the test set, which shows the effectiveness of using CLIP image embeddings over directly predicting the verb from the images.
We compare ClipSitu XTF to the only other CLIP-based situation recognition approach -- CLIP-Event \cite{li2022clip}.
ClipSitu XTF outperforms CLIP-Event on the tasks of verb prediction by 13\%, semantic role labeling (value) by 14\%, and  noun localization (top-1 predicted verb -- grnd value) by 18\%.
ClipSitu XTF also performs the best for noun prediction based on both the predicted top-1 verb and top-5 verb for value and value-all matrices.
ClipSitu XTF betters state-of-the-art CoFormer by a massive margin of 14.1\% on top-1 value and by 9.6\% on top-1 value-all using the top-1 predicted verb on the test set. 
For noun localization, ClipSitu XTF uses \textit{cross-attention scores} to obtain better performance than state-of-the-art CoFormer \cite{cho2022collaborative} that uses verb and role information only.
ClipSitu XTF shows a massive 11\% improvement in noun localization over CoFormer When the ground-truth verb is unavailable (grnd value for top-1 predicted verb).
We explain this performance improvement with a detailed comparison of cross-attention scores to verb-role embedding in \cref{tab:grndcompare}.

\subsection{Video Situation Recognition}
\label{sec:vidsitu_sota}
\begin{table*}
\centering
\scriptsize
\begin{tabular}{l|l|c|c|c|c|c|c} 
\hline
\multicolumn{2}{l|}{Method} & \begin{tabular}[c]{@{}c@{}}Object \\BBoxes\end{tabular} & CIDEr & C-Vb & C-Arg & R-L & Lea \\ 
\hline
\multicolumn{2}{l|}{OSE \cite{yang2023video}}  & Yes & 48.46 & 56.04 & 44.60 & 41.89 & - \\ 
\multicolumn{2}{l|}{{VideoWhisperer \cite{khan2022grounded}}}  & Yes & \textbf{{68.54}} & \textbf{{77.48}} & \textbf{{61.55}} & \textbf{{45.70}} & {47.54} \\ 
\hline
\multicolumn{2}{l|}{SlowFast+TxE+TxD \cite{sadhu2021visual}} 
  & No & 46.01 & 56.37 & 43.58 & 43.04 & \underline{50.89} \\ 
\multicolumn{2}{l|}{Slow-D+TxE+TxD \cite{xiao2022hierarchical}}  & No & 60.34 & 69.12 & 53.87 & \underline{43.77} & 46.77 \\ 
\multicolumn{2}{l|}{VideoWhisperer \cite{khan2022grounded}}  & No & 47.91 & - & - & - & -\\ 
\hline 
\multicolumn{2}{l|}{ClipSituMLP+TxD} & No & \underline{61.93} & \underline{70.14} & \underline{56.30} & \underline{43.77} & 37.77 \\ 
\hline
\end{tabular}
\caption{Comparison with state-of-the-art on Video Situation Recognition. TxE -- Transformer Encoder, TxD -- Transformer Decoder, C-Vb -- CIDEr Verb, C-Arg -- CIDEr Argument, R-L -- Rouge-L. ClipSituMLP+TxD performs the best in 4 out of 5 metrics among all approaches that do not use object bounding boxes. \underline{Underline} denotes second best performance.}
\label{tab:vidsitu_sota}
\end{table*}

We compare the ClipSitu models with transformer decoder on the VidSitu dataset in \cref{tab:vidsitu_sota}.
Our results are reported on the validation set following existing approaches.
To compare fairly with existing approaches, we test the ClipSitu models using ground-truth roles  provided for every verb.
Additional information in the form of object bounding boxes is used in \cite{yang2023transforming, khan2022grounded} due to which \cite{khan2022grounded} performs the best among the existing approaches. 
When trained only on videos in the absence of object bounding boxes, overall CIDEr drops by about 10\% for \cite{khan2022grounded} but they do not provide other metrics for this setting.
Our ClipSitu models for VidSitu do not consider object features making them easily extensible from imSitu to VidSitu with the addition of a decoder.
\revise{We wanted ClipSitu to predict situations directly from the video as object bounding boxes are highly dependent on object detection accuracy and ground truth bounding boxes will not be available in practice.
Even when compared to approaches trained with bounding boxes \cite{yang2023transforming, khan2022grounded}, ClipSituMLP+TxD is comparable in performance on Rouge-L.
Among all the metrics, ClipSitu results are lower on Lea suggesting that ClipSitu sometimes misidentify key roles with incorrect nouns, resulting in incomplete or incorrect scene interpretations. 
Lea weights each entity by its "importance" and "resolution score," a low score might suggest that ClipSitu commits ambiguous role assignments such as confusing "driver" with "passenger" weakening the accuracy and coherence of the scene interpretation.
In summary, comparing among approaches that do not use object bounding boxes, ClipSituMLP+TxD performs comparable or better than existing approaches on 4 out of 5 metrics.}

\section{Situational Summary of Out-of-Domain Images}
\label{sec:captioning}
\begin{figure*}
\centering
\includegraphics[width=\linewidth]{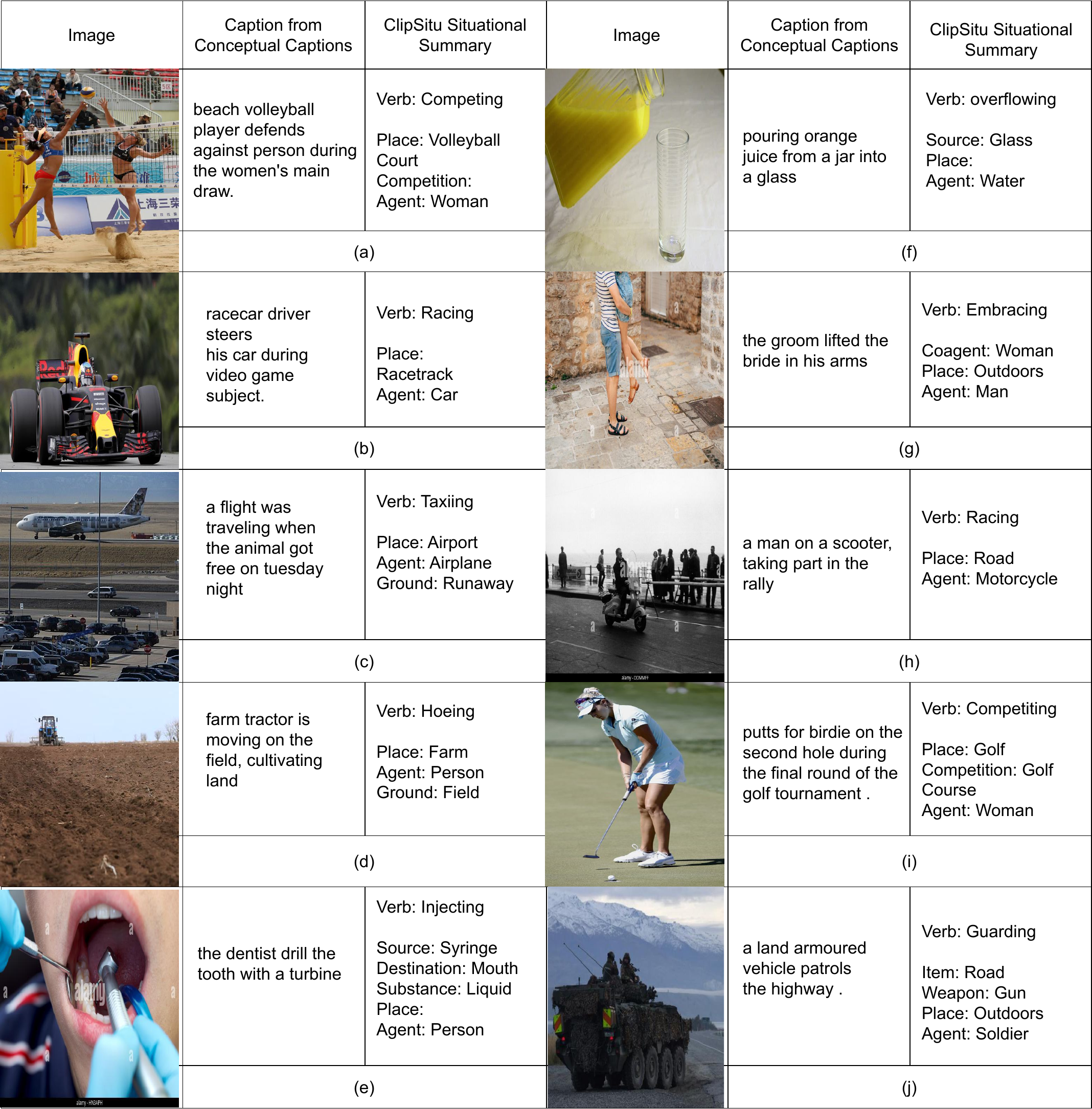}
\caption{Examples of ClipSitu situational summaries on out-of-domain action-oriented images from Conceptual Captions  \cite{sharma2018conceptual}. Compared to the original captions, situational summaries of ClipSitu concisely describes relevant verbs, associated roles and the nouns playing those roles. For certain images, such as (f) ``overflowing'', (h) ``racing'', ClipSitu tries to force a verb even when a faint or no relation to the action exists. In some cases, ClipSitu can return null for some roles such as (e), (f).}
\label{fig:clipsitu_summary}
\end{figure*}
 \label{sec:role_pred}
In existing semantic role labeling approaches, the roles are provided based on the verb to predict the corresponding noun.
For example, as shown in \cref{fig:qual_noun}, for the verb \textit{cramming}, we need to provide the roles \textit{theme, agent,} and \textit{container}. 
This restricts us from building an end-to-end framework that describes any image using situational frames where the ground-truth verb and role annotations are not available.
To overcome this limitation, we introduce a novel role prediction model using the imSitu dataset.

\subsection{Role prediction on imSitu} 
Our role prediction model leverages a multi-headed MLP architecture to simultaneously predict multiple roles associated with a given verb. 
Each head outputs a probability distribution across 191 classes -- 190 for the unique roles and an additional class for cases where a role does not exist.
The input to our model consists of concatenated image and verb embeddings $[X_I;X_V]$.
\revise{As roles are context specific, we cannot predict them without knowing the verb first.
For example, the image in \cref{fig:buy_or_sell} can be perceived as \textit{buying} or \textit{selling}. 
In such a case, the roles are dependent on the verb and we need to use the verb embedding of the predicted verb.
\begin{figure}
    \centering
    \includegraphics[width=0.5\linewidth]{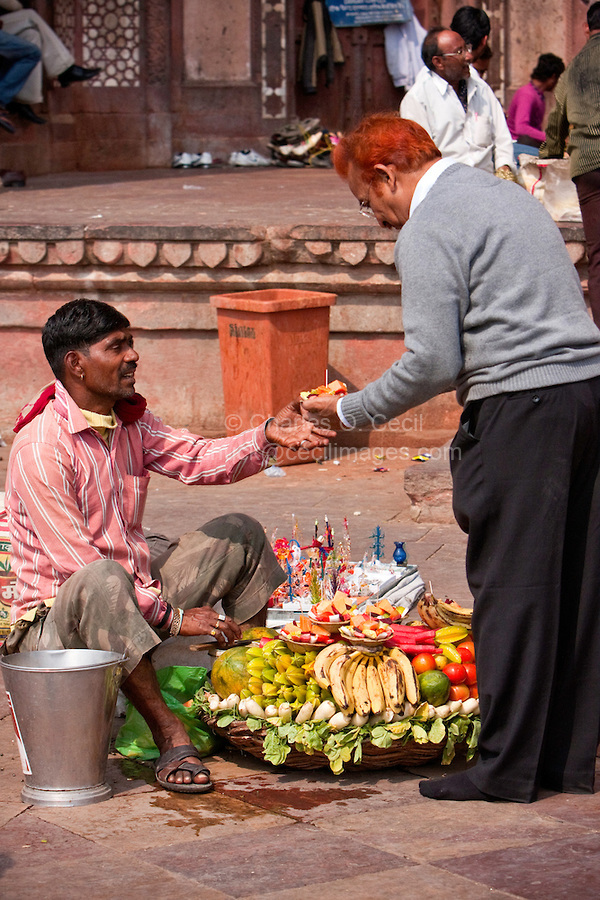}
    \caption{"buying" or "selling"? GT: buying}
    \label{fig:buy_or_sell}
\end{figure} 
}

We utilized various CLIP encoder configurations, from ViT-B/16 and ViT-B/32 to ViT-L/14 and ViT-L/14-336, to generate these embeddings. 
The model output $\hat{r}_{i,j}; i\in \{1,\cdots,m\},j\in\{1,\cdots,R\}$ is the predicted role out of the 191 classes ($R$) for each head. 
As we need to predict a maximum of 6 roles ($m=6$) per input, we trained the model using \textit{multi-head cross-entropy} loss that computes the cross-entropy loss for each prediction head (role) separately and returns the average loss across all heads. 
\begin{align}
    \mathcal{L}_{ROLE} = - \frac{1}{m} \frac{1}{R} \sum_{i=1}^{m}\sum_{j=1}^R r_{i,j} log(\hat{r}_{i,j}).
\end{align}
$\mathcal{L}_{ROLE}$ ensures balanced learning across all the roles of a verb.
Remarkably, our model achieves near-perfect training and validation accuracy, ranging between 99-100\%, within one epoch for all tested CLIP encoder configurations. 
With this exceptional level of accuracy, ClipSitu enables us to predict situational frames directly from images.

\subsection{Situational Summary on Conceptual Captions Dataset}
Our role prediction model completes our end-to-end pipeline of obtaining situational summaries from an image.
Therefore, we present an application of ClipSitu -- situational summaries of out-of-domain images. 
We select images from Conceptual Captions dataset \cite{sharma2018conceptual} that encompass a wide range of visual concepts and abstractness levels.

We show examples from Conceptual Captions in \cref{fig:clipsitu_summary} comparing 
situational summaries from ClipSitu and the captions in Conceptual Captions. 
ClipSitu effectively identifies relevant verbs and roles for action-oriented images, providing a concise representation of the key elements and relationships within an image.
ClipSitu results are more grounded in the image than captions which can be sometimes incorrect \cref{fig:clipsitu_summary}(g) or contain additional information not observed in the image that cannot be verified \cref{fig:clipsitu_summary}(b), (c), (h), (i).
ClipSitu faces challenges with more abstract images or cases where it attempts to force an action or activity even when only a faint relation to the action exists \cref{fig:clipsitu_summary}(f), (h).
\section{Situation Recognition with Large Vision Language Model (VLM) } \label{sec:lvlm}
Large vision language models (VLM) are shown to learn new tasks with a few examples provided in context.
We study the performance of a contemporary state-of-the-art VLM i.e. VILA \cite{Lin_2024_CVPR} on both image and video semantic role labeling when provided in-context examples.
We choose VILA as it has been shown to perform better than other vision language models such as LLaVA \cite{liu2023llava}. 
Furthermore, it allows in-context examples with interleaved image and text inputs. 
We also show the effect of completely finetuning an VLM such as VILA on the tasks of image and video situation recognition and the performance improvement obtained after the process.
\begin{figure}[!htbp]
    \centering
    \includegraphics[width=\linewidth]{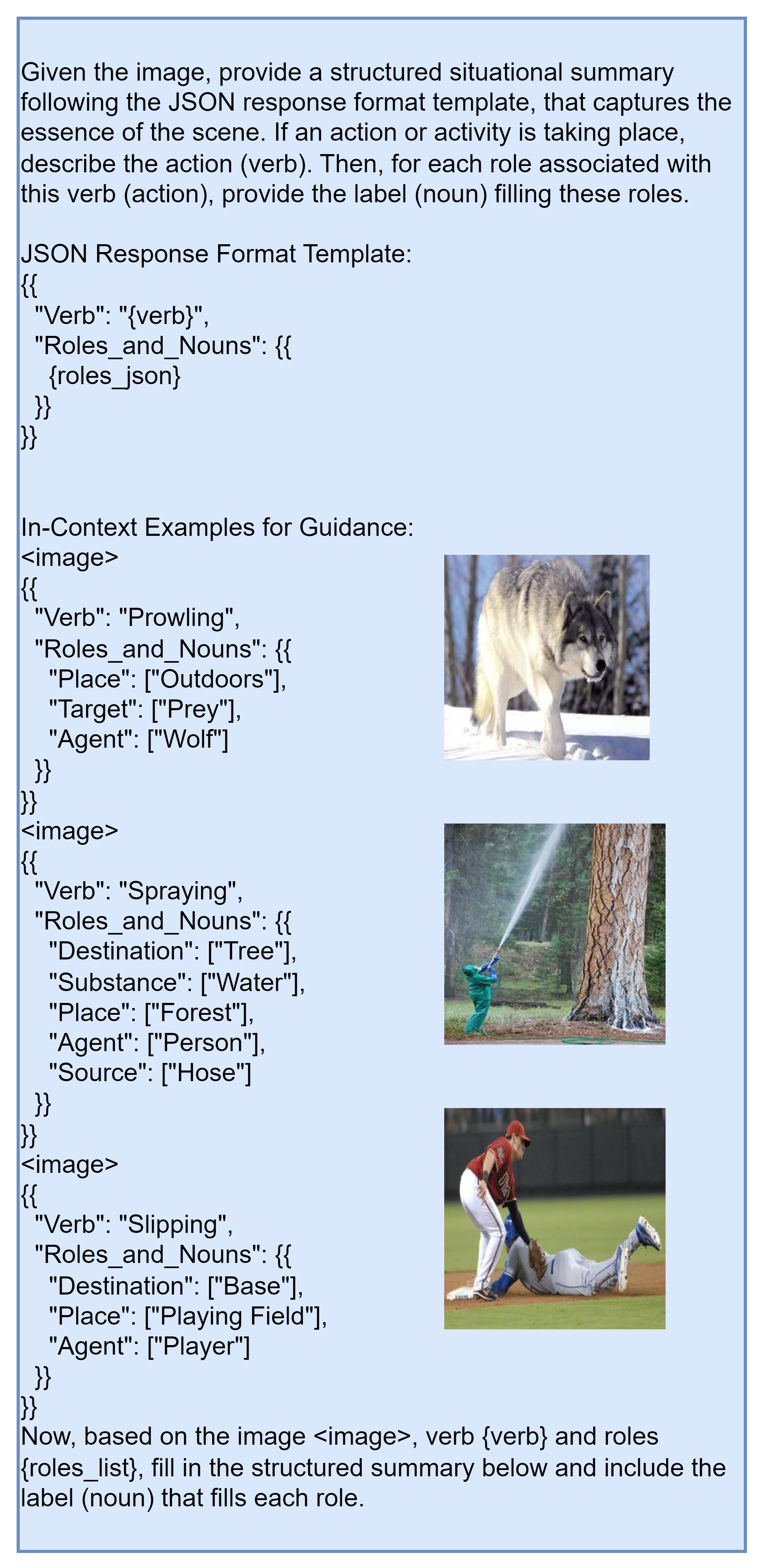}
    \caption{Prompt template with constrained JSON output for each imSitu image with 3 fixed in-context examples. Images are appended after the prompt but shown together with the prompt here for clarity. Best viewed on a screen. }
    \label{fig:prompt}
\end{figure}

\subsection{Image Situation Recognition with VLM in-context}
\begin{figure}
    \centering
    \includegraphics[width=\linewidth]{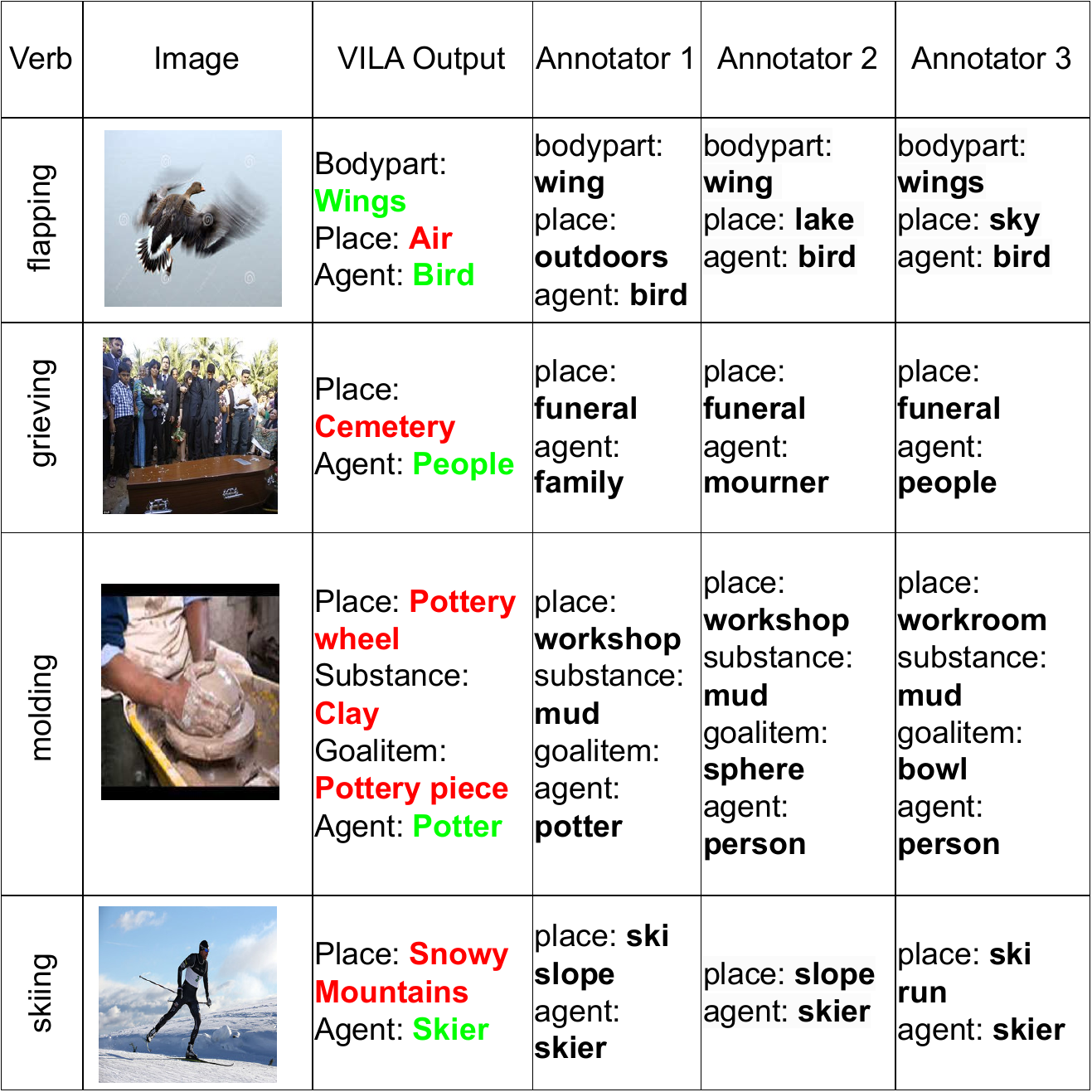}
    \caption{Qualitative examples comparing the generated nouns from VILA with the ground-truth nouns for all 3 annotators in imSitu dataset. VILA generates semantically similar nouns but not the same nouns leading to lower value-all performance. Correct nouns in \textcolor{green}{green} while incorrect are shown in \textcolor{red}{red}.}
    \label{fig:qual_vila}
\end{figure}

The prompt sent to VILA with the in-context examples with images as shown in \cref{fig:prompt}.
We fix these 3 in-context examples in the prompt to VILA to generate the nouns across all the imSitu dev and test sets. 
With this limited context, we provide the image, verb and corresponding roles during testing and only ask VILA to generate the nouns associated with each role following the semantic role labeling experiments with grounded verbs. 
As shown in \cref{tab:vila_incontext}, VILA performs decently on value compared to the fine-tuned ClipSitu XTF, but in terms of value-all, performance of VILA is poor.
This is perhaps because VILA generates semantically similar nouns that do not exactly match the ground-truth nouns as shown in \cref{fig:qual_vila}.
Therefore, even if a single generated role from VILA matches the ground-truth, all the predictions do not match the ground-truth from any of the 3 annotators leading to poorer value-all performance.

\subsection{Instruction-tuning VLM for Image Situation Recognition}\label{sec:tuned_vlm_imsitu} 
Predicting all the roles for an image is quite challenging (value-all) for an VLM such as VILA with in-context and requires additional knowledge of the image-verb-role-noun relations in situational description.
To embed this knowledge into VLMs, we fine-tune them using instruction tuning for verb prediction and semantic role labeling.
\revise{For broader comparison across multiple VLMs, we fine-tune 3 state-of-the-art large visual-language models -- LLaVA1.6-mistral-7B \cite{liu2023llava}, MiniCPM-V2.6 \cite{yao2024minicpm}, and VILA1.5-3B, with the image and provide the verb, roles, and noun as the label.}
The image and label is provided to all the VLMs in the following instructional form --

\noindent \texttt{<image>, Task ImSitu: Generate the verb and corresponding roles, then generate the noun based on the verb and roles\\
\{"verb": <verb>, \\ "roles": \{1: <role1>, 2: <role2>, 3: <role3>, ... \}\\ "nouns": \{ <role1>: <noun1>, <role2>: <noun2>, <role3>: <noun3>, ...\}
\}
}

\revise{Here, \texttt{<image>, <verb>, <role\textit{i}>, <noun\textit{k}>} are placeholders replaced by the image, verb, roles, and nouns.
During testing, we only provide the prompt \texttt{<image>, Task ImSitu: Generate the verb and corresponding roles, then generate the noun based on the verb and roles} and ask the VLM to generate the text containing verb, roles, and nouns.
After collecting the verb and nouns from the generated text, we compare the situation recognition  results of fine-tuned VLMs with ClipSitu XTF in \cref{tab:ft_vlm_imsitu}. 
ClipSitu XTF offers reasonable performance on both verb prediction and semantic role labeling with a fraction of the parameters (45.3M) compared to the VLMs.
All fine-tuned VLMs perform only slightly better than ClipSitu XTF at verb prediction which shows that VLMs cannot completely predict the exact situational verb exactly as the human annotators but predicts a semantically similar verb.
On single role prediction, all the VLMs perform almost twice as well as XTF demonstrating the prowess of VLMs at identifying at least a single role correctly.
However, predicting the nouns for all roles of a verb correctly is still challenging for VLMs as shown by the performance on value-all. 
VLMs do not drastically outperform ClipSituXTF for verb prediction and value-all as they generate verbs and nouns that are semantically similar to the ground-truth nouns for each role (as shown for VILA in \cref{fig:qual_vila}) but not the exact ground-truth nouns which leads to low value-all performance.
Therefore, it is difficult to justifiably judge VLM performance on the value-all metric that is not flexible to account for semantic similarity.
}

\begin{table}[]
    \centering
    \resizebox{\linewidth}{!}{
    \begin{tabular}{ccc|cc}
\hline
\multicolumn{1}{c|}{\multirow{2}{*}{Method}} & \multicolumn{2}{c|}{dev} & \multicolumn{2}{c}{test} \\
\multicolumn{1}{c|}{} & value & \multicolumn{1}{c|}{value-all} & value & value-all \\ \hline
\multicolumn{1}{c|}{\begin{tabular}[c]{@{}c@{}}VILA \cite{Lin_2024_CVPR}\\(in-context) \end{tabular}} & 67.44 & 05.03 & 67.30 & 05.17 \\ \hline
\multicolumn{1}{c|}{\begin{tabular}[c]{@{}c@{}}ClipSitu XTF\end{tabular}} & \textbf{78.52} & \textbf{45.31} & \textbf{78.52} & \textbf{45.31} \\ \hline
\end{tabular}
}
\caption{Comparison of large vision language model (VILA) with in-context examples to ClipSitu XTF on image semantic role labeling in imSitu.
}
\label{tab:vila_incontext}
\end{table}

\begin{table}
\centering
\scriptsize
\resizebox{\linewidth}{!}{
\begin{tabular}{l|ccc} 
\hline
\multirow{3}{*}{Method} & \multicolumn{3}{c}{Top-1 predicted verb} \\ 
\hhline{~---}
 & \multicolumn{3}{c}{dev/test}  \\ 
\hhline{~---}
 & \multicolumn{1}{l}{verb} & \multicolumn{1}{l}{value} & \multicolumn{1}{l}{value-all} \\ 
\hline
\begin{tabular}[c]{@{}l@{}} \revise{LLaVA} \cite{liu2023llava} \\\revise{(instruction-tuned)}\end{tabular} & \revise{58.77/59.25} & \revise{94.29/94.43} & \revise{\textbf{30.63/30.61}}\\ 
\begin{tabular}[c]{@{}l@{}} \revise{MiniCPM-V} \cite{yao2024minicpm} \\ \revise{(instruction-tuned)}\end{tabular} & \revise{57.37/58.04} & \revise{94.93/95.02} & \revise{30.02/30.36} \\ 
\begin{tabular}[l]{@{}c@{}}VILA \cite{Lin_2024_CVPR}\\(instruction-tuned) \end{tabular} & \textbf{58.90}/\textbf{59.32} & \textbf{95.00}/\textbf{95.11} & 26.36/26.44 \\
\hline
ClipSituXTF & 58.19/58.09 & 47.23/47.21 & 29.73/29.61 \\
\hline
\end{tabular}
}
\caption{Comparing instruction-tuned VLMs to ClipSitu XTF on image situation recognition in imSitu.} 
\label{tab:ft_vlm_imsitu}
\end{table}

\subsection{Video SRL with VLM in-context}
We also test the performance of large vision language model VILA \cite{Lin_2024_CVPR} on VidSitu for semantic role labeling of video events.
As VILA accepts only images as in-context examples, we use the strategy of using 2 representative images per event (extracted at 1 frame per second for every 2 second event) as adopted for object extraction in \cite{khan2022grounded}. 
Similar to imSitu, we provide the image, verb and roles to VILA and ask it to generate role-wise nouns.
From every image belonging to an event, we aggregate a list of nouns for every role.
We compute the generation metrics in \cref{tab:vila_vidsitu_incontext} by comparing the aggregated list with the ground-truth annotations of 3 annotators per event in VidSitu \cite{sadhu2021visual}.
The CIDEr metrics of VILA with in-context examples in \cref{tab:vila_vidsitu_incontext} show that it is not able to generalize its understanding of the variety of verbs and their corresponding roles as well as a fine-tuned ClipSitu MLP+TxD model.
Another reason is that the events are temporal in nature and providing 2 frames per event might not be sufficient to visually describe the event.

\begin{table}[]
\centering
\resizebox{0.6\linewidth}{!}{
\begin{tabular}{c|c}
\hline
{Method} & {CIDEr-Arg} \\ \hline
\begin{tabular}[c]{@{}c@{}}VILA \cite{Lin_2024_CVPR}\\(in-context) \end{tabular}& 31.97 \\
\hline
\begin{tabular}[c]{@{}c@{}}ClipSituMLP+TxD\end{tabular} & \textbf{56.30} \\
\hline
\end{tabular}
}
\caption{Comparing VILA with in-context examples versus ClipSituMLP+TxD on Video Semantic Role Labeling in VidSitu.}
\label{tab:vila_vidsitu_incontext}
\end{table}

\subsection{Instruction-tuning VLMs for Video Situation Recognition}\label{sec:tuned_vlm_vidsitu} 
\revise{To overcome these limitations of in-context examples, we opt to fine-tune 4 state-of-the-art VLMs (Video LLMs) for the video situation recognition task (verb prediction and semantic role labeling) -- VILA \cite{Lin_2024_CVPR}, MiniCPM-V \cite{yao2024minicpm}, Qwen2-VL \cite{wang2024qwen2}, and LLaVA-Video \cite{zhang2024video}.
Specifically, we fine-tune  the VILA1.5-3B, MiniCPM-V2.6, Qwen2-VL-7B, LLaVA-Video7B models using instructions by providing the video, verb, roles, and nouns.}
Our instruction for the video situation recognition task is as follows --

\noindent \texttt{<video>, Task VidSitu: Generate the verb and corresponding roles, then generate the noun based on the verb and roles\\
\{"verb": <verb>, \\ "roles": \{1: <role1>, 2: <role2>, 3: <role3>, ... \}\\ "nouns": \{ <role1>: <noun1>, <role2>: <noun2>, <role3>: <noun3>, ...\}
\}
}
\begin{table}
\centering
\scriptsize
\resizebox{\linewidth}{!}{
\begin{tabular}{l|c|c|c|c} 
\hline
Method & CIDEr & CIDEr-Vb & CIDEr-Arg & Rouge-L \\ 
\hline
\begin{tabular}[c]{@{}l@{}}VILA \cite{Lin_2024_CVPR} \\(instruction-tuned)\end{tabular} & 40.44 & 81.68 & 65.07 & 41.80 \\
\begin{tabular}[c]{@{}l@{}}\revise{MiniCPM-V} \cite{yao2024minicpm} \\\revise{(instruction-tuned)} \end{tabular} & \revise{53.04} & \revise{113.44} & \revise{77.53} & \revise{42.33} \\ 
\begin{tabular}[c]{@{}l@{}} \revise{Qwen2-VL}\cite{wang2024qwen2} \\\revise{(instruction-tuned)}\end{tabular} & \revise{57.28} & \revise{127.34} & \revise{76.92} & \revise{41.58}\\ \
\begin{tabular}[c]{@{}l@{}} \revise{LLaVA-Video} \cite{zhang2024video} \\\revise{(instruction-tuned)}\end{tabular} & \revise{60.10} & \revise{\textbf{133.77}} & \revise{\textbf{78.04}} & \revise{42.81}\\
\hline
ClipSituMLP+TxD & \textbf{61.93} & 70.14 & 56.30 & \textbf{43.77} \\
\hline
\end{tabular}
}
\caption{Comparing instruction-tuned VLMs (Video LLMs) to ClipSituMLP+TxD on video situation recognition (VidSitu).}
\label{tab:ft_vila_vidsitu}
\end{table}
Here,\texttt{<video>, <verb>, <role\textit{i}>, <noun\textit{k}>} are placeholders replaced by the video, verb, roles, and nouns.
Similar to imSitu finetuning, we only provide the prompt \texttt{<video>, Task VidSitu: Generate the verb and corresponding roles, then generate the noun based on the verb and roles} and ask the VLM to generate the text containing verb, roles, and nouns.
\revise{After collecting the verb and nouns from the generated text, we compare the situation recognition results of instruction-tuned video LLMs with our best model from \cref{tab:vidsitu_sota} - ClipSituMLP+TxD in \cref{tab:ft_vila_vidsitu}.
ClipSituMLP+TxD being a lightweight model performs quite well on all the metrics compared to existing models but video LLMs with world knowledge and a large number of parameters are able to better describe the verb and nouns on the video.
Strangely, on overall CIDEr and Rouge-L, VideoLLMs do not perform as well.
Overall CiDEr in VidSitu is evaluated by inserting the predicted verbs and nouns in the template set in \cite{sadhu2021visual} i.e. ``verb <verb name> Arg0 <noun1 name> Arg1 < noun2 name> ...`` and matching with the ground truth sentence.
CIDEr is designed to be evaluated with multiple reference sentences but here we only have a single ground truth sentence.
When evaluating long sentences, deviation of a single predicted noun from the ground truth noun is severely penalized which leads to lower overall Rouge-L.
Therefore, overall CIDEr and Rouge-L is not a reliable metric for evaluating the video LLMs and CIDEr-Vb and CIDEr-Arg should be considered as the true performance of the video LLMs.
}

\section{Conclusion}
In conclusion, our work on ClipSitu represents a significant advancement in situation recognition for both images and videos. 
By leveraging the powerful capabilities of CLIP-based image embeddings, we have developed an end-to-end framework that excels in predicting semantic roles associated with verbs, thereby generating comprehensive situational summaries even for out of domain images. 
Our approach enhances the accuracy and contextual relevance of the generated situational summaries of images and videos as shown through extensive experimentation.
Furthermore, ClipSitu's integration of a transformer decoder to generate nouns as text phrases from video frames highlights its versatility and robustness. This extension is crucial for applications requiring detailed and structured descriptions of dynamic scenes, such as multimedia retrieval and event monitoring. 
Our results demonstrate the potential of ClipSitu to serve as a foundational tool in visual media understanding, offering rich, action-centric descriptions that can be evaluated for correctness and integrated with external knowledge bases for enhanced reasoning and decision-making.

In future, there are several promising directions for future research. One area of interest is the enhancement of ClipSitu's ability to handle more complex and nuanced interactions within images and videos, potentially through the incorporation of additional contextual information and multimodal data. 
Another promising directions is exploring the application of ClipSitu in specialized domains such as medical imaging or autonomous driving could yield valuable insights and improvements in those fields. 
Finally, we aim to refine the model's scalability and efficiency, ensuring it can be effectively deployed in real-world scenarios with large-scale datasets and diverse visual content.


\textbf{Acknowledgments}
This research/project is supported by the National Research Foundation, Singapore, under its NRF Fellowship (Award\# NRF-NRFF14-2022-0001). This research is also supported by funding allocation to Basura Fernando by the Agency for Science, Technology and Research (A*STAR) under its SERC Central Research Fund (CRF), as well as its Centre for Frontier AI Research (CFAR).

\bibliographystyle{unsrt}
\bibliography{references}

\end{document}